\documentclass[11pt]{article}

\PassOptionsToPackage{table,xcdraw}{xcolor}
\usepackage[final]{acl}
\usepackage{times}
\usepackage{latexsym}
\usepackage{float}
\usepackage[T1]{fontenc}
\usepackage[utf8]{inputenc}
\DeclareUnicodeCharacter{200B}{}
\usepackage{microtype}

\usepackage{inconsolata}

\usepackage{graphicx}

\usepackage{multicol}
\usepackage{multirow}
\usepackage{amsfonts,amsmath,amssymb}
\usepackage{tcolorbox}
\tcbuselibrary{breakable}
\usepackage{xcolor}
\usepackage{enumitem}
\usepackage{microtype}
\usepackage{array}
\usepackage{longtable}
\usepackage{booktabs}
\usepackage{url}
\usepackage[table,xcdraw]{xcolor}
\usepackage{xpinyin}
\newcommand{\chinese}[1]{%
  \begin{CJK}{UTF8}{gbsn}#1\end{CJK}%
}

\definecolor{headerblue}{RGB}{30,70,150}
\definecolor{pastelteal}{RGB}{100, 200, 190}

\title{ReViCo: Unveiling the Limitations of VLMs in Visual Text Understanding via Error Correction} %Across 22 Nations}

\author{
  Bojun Zhang\textsuperscript{1,2},
  Junhong Liang\textsuperscript{4},
  Feifei Zhai\textsuperscript{1}\thanks{Corresponding author.},
  Fengxian Ji\textsuperscript{4},
  Yu Zhou\textsuperscript{1,3} \\
  $^1$ State Key Laboratory of Multimodal Artificial Intelligence Systems, \\
  Institute of Automation, CAS, Beijing, China \\
  $^2$ School of Artificial Intelligence,
  University of Chinese Academy of Sciences, Beijing, China \\
  $^3$ Fanyu AI Laboratory,
  Zhongke Fanyu Technology Co., Ltd, Beijing, China \\
  $^4$ Mohamed bin Zayed University of Artificial Intelligence \\
  \texttt{\{zhangbojun2022,feifei.zhai\}@ia.ac.cn}, \texttt{yzhou@nlpr.ia.ac.cn} \\
    \texttt{\{junhong.liang,fengxianj.ji\}@mbzuai.ac.ae} \\
}
\begin{document}
\maketitle
\begin{abstract}

Vision Language Models (VLMs) have shown great success in general visual tasks, yet they still struggle to deeply understand text within images. In this paper, we introduce \textbf{ReViCo} (\textbf{Re}al \textbf{Vi}sual \textbf{Co}rrection), a benchmark designed to evaluate VLM text understanding through a novel task of \textit{visual text error correction}. ReViCo challenges models to identify and fix text errors in real-world images, which requires a profound understanding of the interplay between visual text and its surrounding visual context. We benchmark various VLMs using two distinct paradigms: prompt-based strategy and targeted model training, both aimed at pushing the limits of current models. Our experiments reveal a striking performance gap between even the best VLMs and human, and further analysis also shows that most models struggle to accurately perceive the visual text, resulting in frequent correction errors. By highlighting these gaps, ReViCo provides a new benchmark foundation for developing more robust and text-aware VLMs.

\end{abstract}

\section{Introduction}
\label{sec:introduction}

Recent advancements in Vision-Language Models (VLMs) have significantly improved performance on multimodal tasks such as visual question answering (VQA)~\cite{jian-etal-2024-large,guo-etal-2025-crop}, spatial reasoning~\cite{zhang2025mllmsstrugglespatialunderstanding}, and optical character recognition (OCR)~\cite{nagaonkar2025benchmarkingvisionlanguagemodelsoptical}. While these models excel at processing and interpreting general visual inputs, a critical aspect remains underexplored: the depth to which VLMs truly understand and interact with textual content embedded within images.

\begin{figure}
    \centering
    \includegraphics[width=\linewidth]{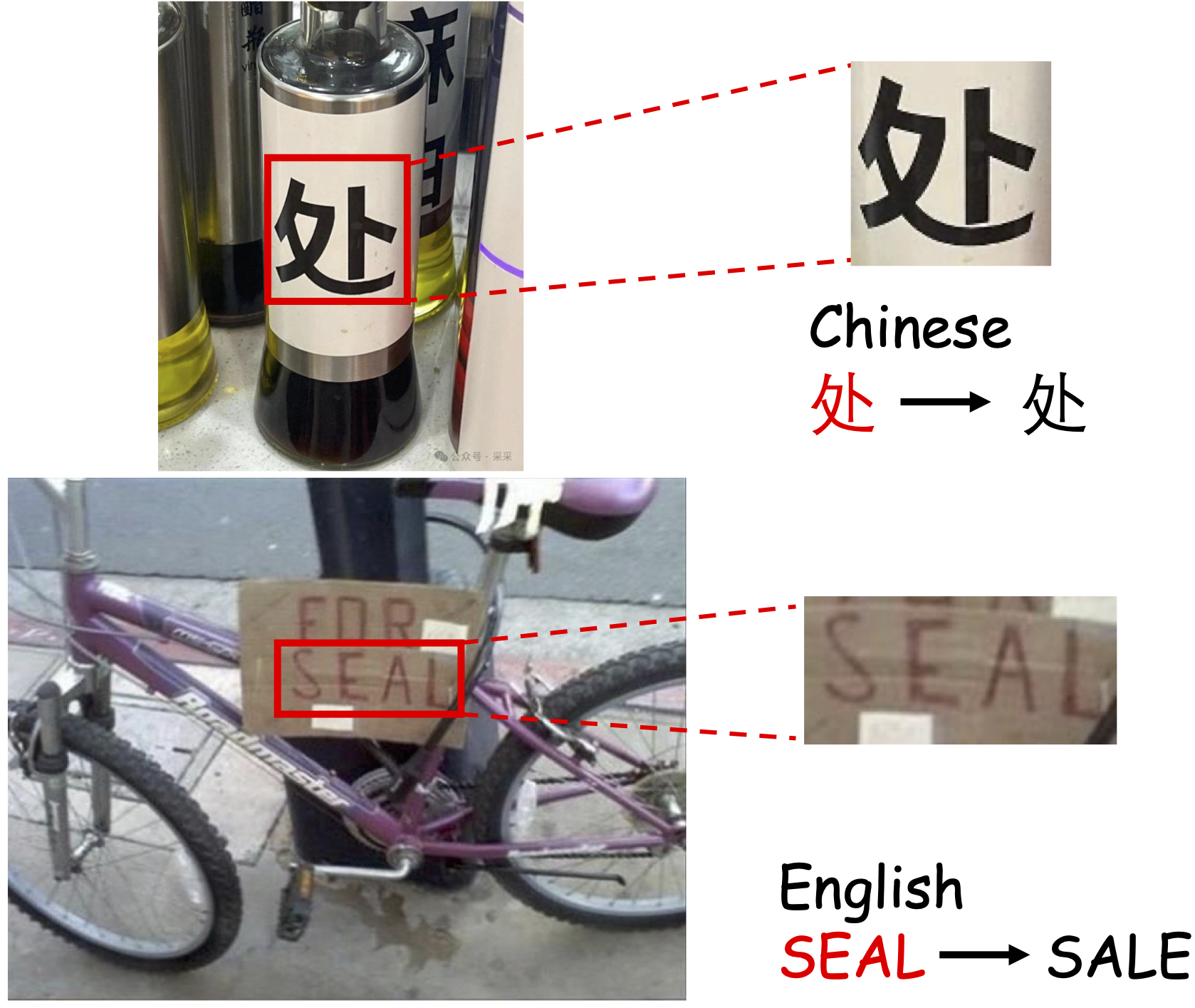}
    \caption{Examples of visual text error correction. Only when a model truly understands the textual content within an image can it perform the correction accurately, both for the Chinese and English example.}
    \label{fig:new_examples_intro}
    \vspace{-1em}
\end{figure}

To evaluate this capability, we approach visual text understanding through a novel task of \textit{visual text error correction} (VTEC). Unlike traditional text error correction, VTEC requires models to deeply understand text within its visual environment, relying on the joint integration of fine-grained visual features and linguistic context. As illustrated in Figure~\ref{fig:new_examples_intro}, a model must grasp the mismatched semantic between the scene(several seasoning bottles) and the text(\chinese{处}) and recognize subtle phonetic similarities between characters (e.g., \chinese{\textcolor{red}{处}} $\rightarrow$ \chinese{\textcolor{blue}{醋}}) for Chinese text correction, while in the English example, it must understand the context(A bike with a cardboard on it) and resolve typographical similarities (e.g., \textcolor{red}{SEAL} $\rightarrow$ \textcolor{blue}{SALE}). Therefore, VTEC serves as an ideal proxy for measuring a VLM's visual text understanding. Success in this task indicates that a model does more than just identify characters — it actually understands the meaning and context of the text it sees.

Driven by this motivation, we introduce \textbf{ReViCo}, a new benchmark designed to assess VLMs' capabilities in comprehensively understanding visual contexts and embedded text via the VTEC task. ReViCo consists of high-quality Chinese and English images featuring natural text errors captured in everyday contexts. Specifically, ReViCo encompasses two key sub-tasks: (1) Error Detection, which requires precisely identifying incorrect tokens/images, and (2) Error Correction, which focuses on restoring them to their correct forms.

On ReViCo, We further perform a comprehensive evaluation of a wide range of representative VLMs, including open-source models (e.g., Qwen-VL, InternVL) and state-of-the-art closed-source systems (e.g., GPT-4o, Gemini2.5-pro and Seed-1.8). To elicit the best performance from these models, we employ two distinct paradigms: prompt-based strategies—including Direct Correction, and Background Information Enhancement to provide different levels of support to models—and targeted model training, which enhances the model's performance on this benchmark by leveraging automatically constructed training data and RL-based (Reinforcement Learning) training methodologies.

Final experimental results reveal a significant performance gap between current VLMs and human-level evaluation. Even with prompt enhancement and targeted training, which do lead to some improvements, the overall performance remains limited. These findings highlight a critical bottleneck: current VLMs have yet to achieve the fine-grained visual text understanding necessary to bridge this performance gap, and our ReViCo actually provides a new benchmark foundation for developing more robust and text-aware VLMs.

Our contributions are summarized as follows:

\begin{enumerate}
    \item We define a new task of visual text error correction and introduce \textbf{ReViCo}, a novel benchmark that provides a new perspective for evaluating VLMs' understanding of visual text through real-world images.
    \item We propose two different paradigms to elicit VLM capabilities, including two prompting strategies and a targeted model training approach to enhance the model’s fine-grained perception of visual text information.
    \item Our experiments and analysis of diverse VLMs highlights the persistent gap between current VLMs and humans, offering valuable insights and benchmark foundation for building more robust and text-aware VLMs.
\end{enumerate}

\section{Related Works}
\label{sec:related_works}

\paragraph{Architecture of VLMs}
The architectures of modern VLMs are primarily designed to bridge the gap between high-dimensional visual signals and discrete linguistic tokens. Many representative models, such as Qwen2.5-VL~\cite{bai2025qwen25vltechnicalreport} and Qwen3-VL~\cite{bai2025qwen3vltechnicalreport}, leverage pretrained vision encoders (e.g., CLIP or ViT) to extract visual features, which are subsequently projected into the embedding space of a pretrained LLM. This approach facilitates the reuse of powerful unimodal foundations but necessitates precise alignment to prevent the loss of fine-grained perceptual information. Other state-of-the-art architectures, like Gemini2.5~\cite{comanici2025gemini25pushingfrontier} and Kimi2.5~\cite{kimiteam2026kimik25visualagentic}, emphasize early text-vision fusion pretraining to enhance reasoning capabilities across modalities, which are recognized as native multimodal models. 

\paragraph{Text Error Correction}
We mainly investigate spelling error correction, which has been extensively studied in both Chinese and English in the text modality. In Chinese, spelling errors primarily arise from phonetic and graphical similarities between characters~\cite{cheng-etal-2020-spellgcn,liu-etal-2021-plome,zhang-etal-2020-spelling}, while in English, they typically result from character-level substitutions or deletions~\cite{electronics9101670,mitton1980birkbeck}. Most existing datasets for spelling correction are derived from purely textual sources, including those from language learners~\cite{tseng-etal-2015-introduction}, native speakers~\cite{hu-etal-2024-cscd}, and specific domains~\cite{liang2025rairretrievalaugmentediterativerefinement}. Methods for addressing text error correction generally fall into two main paradigms: PLM-based approaches, which rely on architectures like BERT~\cite{zhang-etal-2020-spelling}, and LLM-based methods, which often require minimal or no task-specific training and combine PLMs and LLMs to enhance both domain precision and generalization~\cite{qiao2025mixturesmalllargemodels}.

% \paragraph{Text Related VQA}
\paragraph{Reinforcement Learning for Visual Reasoning}
Reinforcement learning (RL), particularly RL from Human Feedback (RLHF)~\cite{christiano2023deepreinforcementlearninghuman,ouyang2022traininglanguagemodelsfollow} and PPO optimization~\cite{schulman2017proximalpolicyoptimizationalgorithms}, has enhanced the reasoning capabilities of LLMs. Recent methods like DPO and KTO optimize preferences without explicit rewards~\cite{rafailov2024directpreferenceoptimizationlanguage,ethayarajh2024ktomodelalignmentprospect}. RL has proven effective for systematic reasoning in tasks like mathematics and logical inference~\cite{cobbe2021trainingverifierssolvemath,lightman2023letsverifystepstep}. The integration of RL in VLMs, such as Qwen3-VL~\cite{bai2025qwen3vltechnicalreport}, improves visual reasoning and spatial understanding. Group Relative Policy Optimization (GRPO)~\cite{shao2024deepseekmathpushinglimitsmathematical} reduces memory overhead and enhances RL scalability in VLMs. Beyond conventional language-space reasoning, recent studies have explored alternative visual reasoning paradigms. Latent Visual Reasoning~\cite{li2025latentvisualreasoning,guo2026visualmemorymechanisticdiagnostics} enables VLMs to perform intermediate reasoning directly in the visual embedding space, whereas tool-augmented reinforcement learning trains models to actively invoke visual tools to acquire, manipulate, and verify task-relevant evidence~\cite{zhang2025thymethinkimages}. These complementary directions broaden the scope of VLM reasoning beyond text-based chains of thought and provide new perspectives on integrating perception and reasoning. 

\begin{figure*}[t!]
  \includegraphics[width=\linewidth]{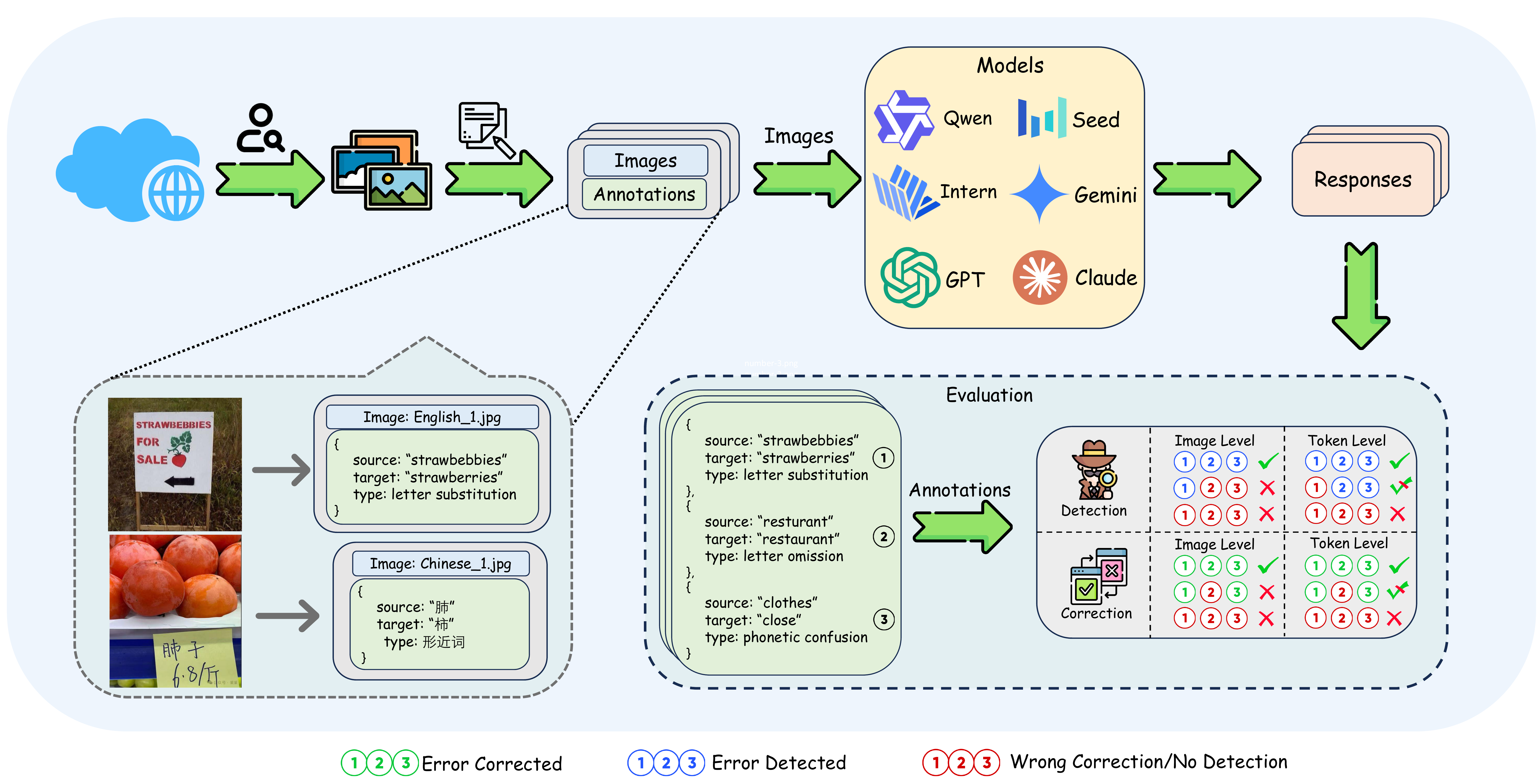}
  \caption{Overview of the ReViCo Benchmark pipeline, illustrated with one English and one Chinese example from the dataset. The evaluation procedure is shown with a single English example.}
  \label{fig:framework}
  \vspace{-4mm}
\end{figure*}

\section{ReViCo Benchmark}
\label{sec:visualcsc_benchmark}
% Bojun did statistics of the dataset
To systematically analyze the visual text error correction ability of LLMs, we propose \textbf{Re}al \textbf{Vi}sual \textbf{Co}rrection (ReViCo) Benchmark. In this section, we introduce the data construction pipeline of both test set and traning set(Sec.~\ref{subsec:construction_pipeline}), the error type statistics of our proposed benchmark(Sec.~\ref{subsec:error_type_stat}) and evaluation protocol(Sec.~\ref{subsec:evaluation_protocol}). An overview of the data gathering and evaluation process is shown in Figure~\ref{fig:framework}.

% \subsection{Task Design}
% \label{subsec:task_design}

% We design two tasks to evaluate the visual text error correction ability of VLMs: \emph{Visual Text Error Detection} and \emph{Visual Text Error Correction}. The first task requires the model to identify if an image contains text errors, such as incorrect characters in Chinese or misspelled/misused words in English. The second task involves both detecting and correcting the errors, where the model must identify erroneous tokens and replace them with the correct ones.

\subsection{Data Construction}
\label{subsec:construction_pipeline}
To ensure both high data quality and construction efficiency, we employ bifurcated strategies for data preparation: manual annotation for the test set to guarantee precision, and automated synthesis for the training set to ensure scalability.

\paragraph{Test Data Construction}
We manually curated the ReViCo test set by collecting images from diverse online resources. To ensure the proposed benchmark evaluates the interplay between visual context and text, annotators were strictly required to select images where errors are only resolvable via visual cues. To evaluate the error detection ability of VLMs, we also sample error-free Chinese from EST-VQA~\cite{wang2020generalvalueevidencebilingual} test set and English images from the EST-VQA and TextVQA~\cite{singh2019towards} test set. The collected error-containing images are randomly divided into two subsets. Each subset is annotated by one volunteer and cross-checked by another to identify misspelled tokens and provide the correct forms. In case of disagreement, a third volunteer acts as an adjudicator to determine the final label. The error-free images are also verified by human annotators. This multi-stage annotation and verification process results in a high-quality real-life labeled dataset for visual text error correction. Finally, we obtain 1229 high-quality, real-world samples.
\paragraph{Automated Training Data Construction}
\label{subsec:reinforcement_learning_data}
We construct a high-quality training dataset for reinforcement learning derived from the EST-VQA~\cite{wang2020generalvalueevidencebilingual} and ST-VQA~\cite{biten2019scenetextvisualquestion} traning split, generated through a three-stage pipeline: \textit{Generate-Edit-Examination}. In the first stage(\textit{Generate}), images from the gathered training set are processed using GPT-4o to create erroneous text by introducing textual distortions. To ensure task difficulty and consistency with the proposed benchmark, a large language model (Qwen3-235B ) filters out samples rectifiable through pure linguistic context. In the second stage(\textit{Edit}), we utilized Qwen-Image-Edit-2509 model to modify the original text regions in the images with the generated erroneous text, resulting in images with textual errors. Finally, in the third stage(\textit{Examination}), Qwen3-VL-235B-A22B-Instruct is employed for image quality examination by performing error correction on the edited images, filtering out low-quality or incorrectly edited images, ensuring the overall quality of the dataset. This automated workflow yielded 4,531 scalable training samples by leveraging the strengths of image editing models and VLMs.

\subsection{Error Type Statistics}
\label{subsec:error_type_stat}
The ReViCo Benchmark consists of Chinese and English datasets with four and seven subsets, respectively. The Chinese dataset includes the following error types: \textit{phonetic error}, \textit{graphic error}, \textit{phonetic-graphic error}, and \textit{others}. The English dataset, due to structural differences, classifies errors at two levels: at the letter level, errors include \textit{letter omission}, \textit{letter insertion}, \textit{letter substitution}, and \textit{letter transposition}; at the word level, errors are classified as \textit{split-word errors}, \textit{phonetic/graphic confusion}, and \textit{others}. A detailed error type analysis is in Appendix~\ref{apdx:error_type_explanation}.

This hierarchical categorization provides a systematic framework for analyzing error types and contributes to a more precise evaluation of text correction methods. Examples of different types of errors for the test set are presented in Figure~\ref{fig:error_type_examples}.

\begin{figure}
    \centering
    \includegraphics[width=\linewidth]{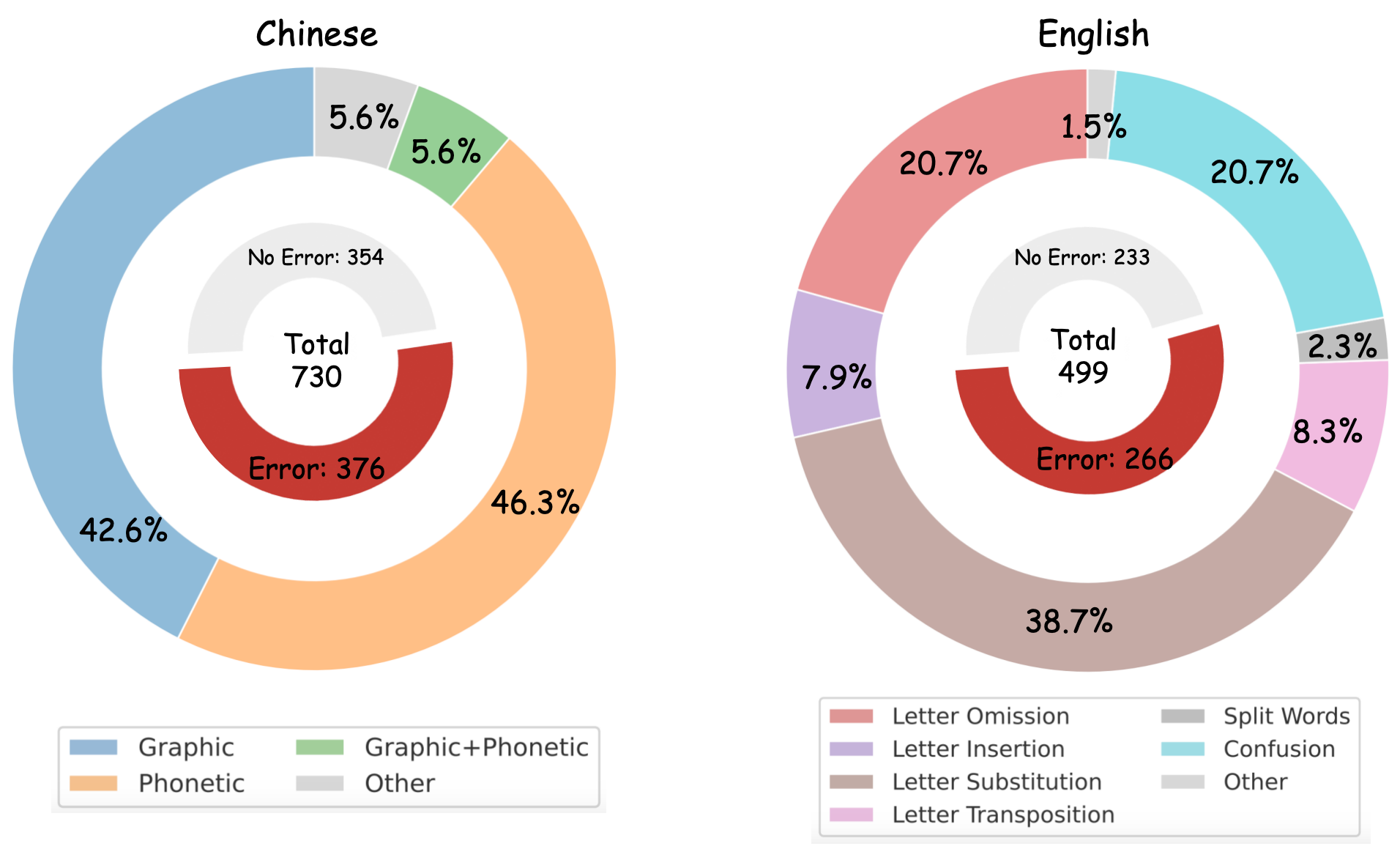}
    \caption{Error statistics of the ReViCo benchmark. For Chinese and English separately, the inner pie charts show the counts of erroneous vs. error-free images, while the outer pie charts depict the distribution of error types across the erroneous images.}
    \label{fig:error_statistics_revico}
    \vspace{-1.5em}
\end{figure}

\begin{table*}[t!]
\centering
\scalebox{0.8}{
\begin{tabular}{lcccccc|cccccc}
\toprule
\multirow{2}{*}{Models} & \multicolumn{3}{c}{\textbf{I}mage-\textbf{D}etection} & \multicolumn{3}{c}{\textbf{T}oken-\textbf{D}etection} 
 & \multicolumn{3}{c}{\textbf{I}mage-\textbf{C}orrection} & \multicolumn{3}{c}{\textbf{T}oken-\textbf{C}orrection} \\ 
\cmidrule(lr){2-4} \cmidrule(lr){5-7} \cmidrule(lr){8-10} \cmidrule(lr){11-13}
 & $P$ & $R$ & $F_1$ & $P$ & $R$ & $F_1$ & $P$ & $R$ & $F_1$ & $P$ & $R$ & $F_1$ \\
\midrule
\multicolumn{12}{c}{\textit{Oracle OCR Text + Large language Models}} \\
\midrule
Qwen3$_{8B}$ 
& 70.7 & 44.1 & 54.3 & 6.7 & 27.5 & 10.8 & 30.3 & 8.0 & 12.6 & 34.3 & 15.4 & 21.3 \\
Qwen3$_{32B}$
& 63.7 & 57.0 & 60.2 & 9.8 & 25.7 & 14.2 & 31.4 & 14.5 & 19.8 & 36.7 & 20.5 & 26.3 \\
Qwen3$_{235B}$
& 77.4 & 58.6 & \underline{66.7} & \underline{24.4} & \underline{33.2} & \underline{28.1} & 55.1 & \underline{21.0} & \underline{30.4} & 45.6 & 24.1 & 30.2 \\
Qwen3VL$_{8B}$ 
& 75.5 & 44.2 & 55.8 & 12.3 & 24.8 & 16.5 & 44.5 & 11.5 & 18.2 & 38.9 & 16.9 & 23.6 \\
Qwen3VL$_{32B}$ 
& \underline{80.0} & 39.7 & 53.1 & 20.2 & 21.9 & 21.0 & \underline{59.1} & 14.3 & 23.1 & \underline{49.0} & 18.9 & 27.3 \\
Qwen3VL$_{235B}$
& 74.6 & \underline{60.2} & 66.6 & 15.4 & 28.5 & 20.0 & 49.6 & 20.2 & 28.7 & 43.3 & \underline{25.8}                  & \underline{31.7} \\
\midrule
\multicolumn{12}{c}{\textit{Image + Vision Language Models}} \\
\midrule
Qwen2.5VL$_{3B}$
& 48.7 & \underline{73.3} & 58.5 & 5.6 & 16.6 & 8.3 & 6.4 & 5.2 & 5.7 & 5.1 & 7.1 & 6.0 \\ 
Qwen2.5VL$_{7B}$ 
& 70.9 & 25.2 & 37.1 & 28.7 & 12.8 & 17.7 & 45.0 & 8.4 & 14.2 & 29.1 & 9.9 & 14.7 \\ 
Qwen2.5VL$_{32B}$ 
& \underline{96.9} & 48.1 & 64.3 & 58.2 & \underline{31.8} & \underline{41.1} & \underline{94.1} & \underline{25.0} & \underline{39.5} & \underline{52.8} & \underline{27.9} & \underline{36.5} \\ 
Qwen2.5VL$_{72B}$
& 93.8 & 49.4 & \underline{64.7} & \underline{59.7} & 31.2 & 41.0 & 87.9 & 23.8 & 37.5 & 50.7 & 24.5 & 33.1 \\

\midrule
Qwen3VL$_{2B}$ 
& 79.2 & 11.4 & 20.0 & 40.2 & 7.4 & 12.5 & 57.1 & 4.0 & 7.5 & 52.0 & 5.9 & 10.6 \\
Qwen3VL$_{4B}$ 
& 55.8 & 50.9 & 53.3 & 13.4 & 29.5 & 18.4 & 25.7 & 13.9 & 18.1 & 34.9 & 16.5 & 22.4 \\
Qwen3VL$_{8B}$ 
& 57.6 & 60.4 & 59.0 & 9.4 & 29.8 & 14.2 & 19.9 & 11.1 & 14.2 & 26.1 & 15.0 & 19.0 \\
Qwen3VL$_{32B}$ 
& \underline{97.3} & 45.7 & 62.2 & \underline{47.1} & 28.6 & \underline{35.6} & \underline{94.2} & 20.3 & 33.4 & \underline{53.9} & 23.4 & 32.7 \\
Qwen3VL$_{235B}$
& 74.7 & \underline{77.2} & \underline{75.9} & 18.9 &  \underline{35.0} & 24.5 & 50.9 & \underline{27.1} & \underline{35.4} & 42.0 &  \underline{30.1} & \underline{35.1} \\
\midrule
InternVL$_{2B}$ 
& 53.8 & 40.1 & 45.9 & 5.8 & 11.7 & 7.8 & 5.8 & 2.1 & 3.1 & 3.3 & 4.0 & 3.7 \\ 
InternVL$_{4B}$
& 55.8 & \underline{67.7} & \underline{61.2} & 5.8 & 14.5 & 8.3 & 8.7 & 5.1 & 6.5 & 5.0 & 6.2 & 5.5 \\ 
InternVL$_{8B}$
& 76.7 & 21.2 & 33.2 & 12.8 & 8.1 & 9.9 & 34.9 & 3.5 & 6.3 & 10.5 & 4.9 & 6.7 \\
InternVL$_{14B}$
& 65.1 & 53.0 & 58.4 & 13.6 & 17.4 & 15.3 & 27.4 & 10.7 & 15.4 & 16.3 & 12.6 & 14.2 \\ 
InternVL$_{38B}$ 
& \underline{89.8} & 33.4 & 48.7 & \underline{48.6} & \underline{18.9} & \underline{27.2} & \underline{79.3} & \underline{14.5} & \underline{24.5} & \underline{46.5} & \underline{16.1} & \underline{23.9} \\
\midrule
GPT\textendash4o   
& \textbf{98.7} & 48.1 & 64.7 & 66.2 & 33.0 & 44.0 & \textbf{97.7} & 26.4 & 41.6 & 59.0 & 30.0 & 38.8 \\
GPT\textendash5.4   
& 82.7 & 69.6 & 75.6 & 35.5 & 35.8 & 35.6 & 65.2 & 27.2 & 38.4 & 48.2 & 31.6 & 38.2 \\
Seed\textendash1.8 
& 80.5 & 92.5 & 86.1 & 48.0 & 69.7 & 56.8 & 70.6 & 53.8 & 61.0 & 69.9 & 64.3 & 67.0 \\
Seed\textendash2.0 
& 85.9 & 95.3 & \textbf{90.4} & 63.8 & \textbf{86.0} & \textbf{73.3} & 82.0 & 71.2 & \textbf{76.2} & \textbf{84.4} & \textbf{83.0} & \textbf{83.7} \\
Gemini2.5\textendash Pro
& 85.9 & 88.7 & 87.3 & \textbf{68.3} & 75.5 & 71.7 & 81.5 & 64.2 & 71.8 & 81.0 & 71.3 & 75.9 \\
Gemini3.1\textendash Pro
& 79.7 & \textbf{97.4} & 87.7 & 63.0 & 82.3 & 71.4 & 74.4 & \textbf{72.0} & 73.2 & 81.4 & 78.9 & 80.1 \\
% Gemini-2.5-pro &  &  &  &  & &  &  &  &  &  &  &  \\
\bottomrule
\end{tabular}
}
\caption{Performance comparison of OCR+LLMs and VLMs on visual text error detection and correction, evaluated at both the image (I) level and token (T) level across detection (D) and correction (C) tasks. The best overall results in each column are highlighted in \textbf{bold}, while the strongest results within the OCR+LLM, QwenVL and InternVL families are \underline{underlined}.Please note that Qwen3VL models are used as LLM in OCR+LLM pipeline(w.o. image input).}
\label{tab:baseline_result_qwen_intern_gpt_new}
\end{table*}

\subsection{Evaluation Protocol}
\label{subsec:evaluation_protocol}
To evaluate the capability of correcting visual text errors, including both Chinese characters and Latin letters, we define two evaluation tasks: \emph{Detection} and \emph{Correction}. For each task, there are two levels: \emph{Image} and \emph{Token}.

\textbf{Detection} focuses on the capacity of the model to recognize the presence of visual text errors. At the \textit{Image Detection} (I-D), the task is to decide whether an image contains any errors at all, treating the problem as a binary classification. In contrast, the \textit{Token Detection} (T-D) moves to a finer granularity, requiring the system to accurately pinpoint the individual erroneous Chinese characters or English words within error-correction pairs. 

\textbf{Correction} evaluates the model's ability to produce correct outputs once errors have been identified. The \textit{Token Correction} (T-C) measures whether the model can transform each erroneous character or word into its proper form, emphasizing precision at the token level. The \textit{Image Correction} (I-C), by comparison, demands a holistic solution: the image is considered successfully corrected only when all errors are resolved without omission or unnecessary alterations. 

We report the precision ($P$), recall ($R$) and $F_1$ scores for each task and for each level. A more detailed description and calculation can be found in Appendix \ref{apdx:calculation_of_metrics}. 

\section{Evaluation Stratetgies}
To provide a comprehensive evaluation, we employ OCR+LLM pipeline(Sec.~\ref{subsec:ocr_llm}), two prompt-based strategies(Sec.~\ref{subsec:two_prompts}), and a targeted model training paradigm(Sec.~\ref{subsec:grpo_training}) to elicit VLM abilities. These methodologies are detailed as follows:
\subsection{OCR+LLM Pipeline}
\label{subsec:ocr_llm}
Although this benchmark is primarily designed to evaluate the capabilities of VLMs, we also provide evaluation results using OCR-extracted text combined with LLMs as a baseline for comparison. Since the accuracy of OCR results depends on the specific tools used, we report performance based on Oracle OCR text processed and annotated by annotators. 
 \subsection{Prompt-based Strategy}
\label{subsec:two_prompts}
\paragraph{Direct Correction Prompt}
\label{subsec:direct_correction_prompt}
Following previous text-centric Visual Question Answering (VQA) tasks\cite{singh2019vqamodelsread,biten2019scenetextvisualquestion,wang2020general,Liu_2024}, we first evaluate our model using a direct prompting baseline, where the model is tasked with generating the final answer explicitly.

\paragraph{Background Information Enhancement Prompt}
\label{subsec:background_info_prompt}
The Background Information Enhancement strategy necessitates that the model generates a descriptive narrative of the visual context prior to error correction; this mechanism facilitates a more effective utilization of global visual cues to guide the visual text correction process. All prompts for evaluation are shown in Appendix~\ref{apdx: prompt_english_chinese} and ~\ref{apdx:prompt_improve}.
\subsection{Targeted Model Training}
\label{subsec:grpo_training}
Despite the efficacy of prompting strategies, we aim to push the boundaries of fine-grained visual text understanding by leveraging automatically constructed datasets via a training-based paradigm. Considering that RL-based(Reinforcement Learning) methods demonstrate superior robustness against forgetting compared to SFT (Supervised Fine-Tuning) approaches~\cite{shenfeld2025rlsrazoronlinereinforcement}, we naturally employ RL to further refine the model's performance.  Specifically, we adopt the Group-based Policy Optimization (GRPO)~\cite{shao2024deepseekmathpushinglimitsmathematical} algorithm to facilitate the emergence of autonomous fine-grained perception within the model.

During training, we sample a group of output sequences $G = \{y_1, y_2, \ldots, y_G\}$ for each input $x$. Each sequence $y_i$ receives a reward $r_i$, which is then normalized within group $G$ to produce advantages $\hat{A}_i$. The token-level optimization objective is formulated as:
\begin{equation}
\resizebox{1.0\hsize}{!}{$
\begin{aligned}
\mathcal{J}_{\text{GRPO}}(\theta) &= -\frac{1}{|G|} \sum_{i=1}^{|G|} \frac{1}{|y_i|} \sum_{t=1}^{|y_i|} \min \left(r_{i,t}(\theta) \hat{A}_{i,t}, \right. \\
& \left. \text{clip} \left( r_{i,t}(\theta), 1 - \epsilon, 1 + \epsilon \right) \hat{A}_{i,t} \right) - \beta D_{\text{KL}}, \\
\end{aligned}
$}
\end{equation}
\begin{equation}
r_{i,t}(\theta)
=
\frac{\pi_\theta(y_{i,t} \mid x, y_{i,<t})}
{\pi_{\text{ref}}(y_{i,t} \mid x, y_{i,<t})}
\end{equation}

\begin{equation}
D_{\text{KL}}
=
D_{\text{KL}}(\pi_\theta \parallel \pi_{\text{ref}})
\end{equation}

where $\epsilon$ and $\beta$ are hyper-parameters.
\paragraph{Multi-Level Verification Reward}
To facilitate effective RL training with GRPO, we propose the \textbf{Multi-Level Verification Reward (MLVR)} system. Rather than using a holistic score, MLVR decomposes the feedback into three hierarchical stages, ensuring that the model masters foundational constraints before optimizing for fine-grained correction. The system consists of the following components:
\begin{itemize}[noitemsep, topsep=2pt]
\item \textbf{Format Constraint:} A binary filter that verifies whether the response adheres to the required structural protocols.
\item \textbf{Image-Level Discrimination Constraint:} A semantic check assessing the model's macro-level judgment of whether the visual text contains errors.
\item \textbf{Token-Level Correction Reward ($R_{corr}$):} A fine-grained metric, such as Edit Distance or Exact Match, that measures the accuracy of the rectified text.
\end{itemize}
% \vspace{-0.05em}
\begin{equation}
\resizebox{0.89\hsize}{!}{$
R_{total} = 
\begin{cases} 
R_{corr}, &\!\!\! \text{if } (x \in \mathcal{D}_{err} \land Err(y)=1) \\
1, &\!\!\! \text{if }(x \in \mathcal{D}_{clean} \land Err(y)=0)\\
0, &\!\!\! \text{only valid format} \\
-1, &\!\!\!  \text{invalid format}
\end{cases}
$}
\end{equation}
where $\mathcal{D}_{err}$ and $\mathcal{D}_{clean}$ denote the sets of erroneous and error-free samples, respectively. We define a binary indicator $Err(y) \in \{0, 1\}$ to signify whether the model detects text errors. This image-level discrimination constraint prevents the model from collapsing into trivial behavioral biases, such as indiscriminate correction (False Positives) or total omission (False Negatives), thereby ensuring robustness across diverse scenarios. Additionally, a format match constraint is implemented to enforce that the final output adheres to the prescribed structure $\mathcal{F}$ (e.g., <answer>[...]</answer>).

We assess the fine-grained quality of answers via the Correctness Reward $R_{corr}$. This reward decomposes evaluation into detection ($\text{Det}$) and correction ($\text{Cor}$) sub-tasks. Furthermore, to mitigate "enumeration strategies" (where the model over-generates corrections to maximize recall), we incorporate a length penalty term. The overall correctness reward is formally defined as follows:

\begin{equation}
\resizebox{0.88\hsize}{!}{$
\begin{split}
R_{corr} =  & \sum_{e \in \mathcal{E}} 0.5 \cdot \big( \mathbb{I}(\text{Det}(e)) + \mathbb{I}(\text{Cor}(e)) \big)  \\
& - \alpha \cdot \max(0, |\hat{\mathcal{E}}| - |\mathcal{E}|)
\end{split}
$}
\end{equation}
where $\mathcal{E}$ and $\hat{\mathcal{E}}$ denote the sets of ground-truth errors and model-predicted corrections, respectively. The term $|\hat{\mathcal{E}}| - |\mathcal{E}|$ penalizes the model for generating redundant corrections (hallucinations), weighted by the hyperparameter $\alpha$, which is set to 0.1 in our experiments. Details on training data construction are available in Sec.~\ref{subsec:reinforcement_learning_data}.

\begin{table*}[!t]
\centering
\setlength{\tabcolsep}{3pt}
\scalebox{0.8}{
\begin{tabular}{lcc|cc|cc|cc}
\toprule
\multirow{3}{*}{\textbf{Model}} & \multicolumn{2}{c}{\textbf{I-D}} & \multicolumn{2}{c}{\textbf{T-D}} & \multicolumn{2}{c}{\textbf{I-C}} & \multicolumn{2}{c}{\textbf{T-C}} \\
 & ${F_{1d}}$ & ${F_{1b}}$ & ${F_{1d}}$ & ${F_{1b}}$ & ${F_{1d}}$ & ${F_{1b}}$ & ${F_{1d}}$ & ${F_{1b}}$ \\
\midrule

Qwen2.5VL$_{3B}$   & 58.5 & 51.1$_{\textcolor{red}{-7.4}}$  & 8.3  & 2.8$_{\textcolor{red}{-5.5}}$ & 5.7  & 0.6$_{\textcolor{red}{-5.1}}$ & 6.0  & 1.5$_{\textcolor{red}{-4.5}}$ \\
\quad $\hookrightarrow$ GRPO & 68.2$_{\textcolor{blue}{+9.7}}$ & 63.4$_{\textcolor{blue}{+4.9}}$ & 9.5$_{\textcolor{blue}{+1.2}}$  & 8.4$_{\textcolor{blue}{+0.1}}$ & 6.9$_{\textcolor{blue}{+1.2}}$  & 5.9$_{\textcolor{blue}{+0.2}}$ & 14.0$_{\textcolor{blue}{+8.0}}$  & 13.9$_{\textcolor{blue}{+7.9}}$ \\

Qwen2.5VL$_{7B}$   & 37.1  & 62.0$_{\textcolor{blue}{+24.9}}$ & 17.7  & 21.6$_{\textcolor{blue}{+3.9}}$ & 14.2  & 20.4$_{\textcolor{blue}{+6.2}}$  & 14.7 & 18.5$_{\textcolor{blue}{+3.8}}$ \\
\quad $\hookrightarrow$ GRPO & 68.7$_{\textcolor{blue}{+31.6}}$  & 73.1$_{\textcolor{blue}{+36.0}}$  & 17.3$_{\textcolor{red}{-0.4}}$ & 27.2$_{\textcolor{blue}{+9.5}}$ & 16.4$_{\textcolor{blue}{+2.2}}$  & 23.3$_{\textcolor{blue}{+9.1}}$ & 18.6$_{\textcolor{blue}{+3.9}}$  & 22.1$_{\textcolor{blue}{+7.4}}$ \\

Qwen2.5VL$_{32B}$  & 64.3  & 66.5$_{\textcolor{blue}{+2.2}}$  & 41.1  & 47.2$_{\textcolor{blue}{+6.1}}$ & 39.5  & 44.5$_{\textcolor{blue}{+5.0}}$ & 36.5  & 43.6$_{\textcolor{blue}{+7.1}}$\\
Qwen2.5VL$_{72B}$  & 64.7 & 67.8$_{\textcolor{blue}{+3.1}}$  & 41.0  &48.6$_{\textcolor{blue}{+7.6}}$   & 37.5  & 46.6$_{\textcolor{blue}{+9.1}}$     & 33.1   &45.8$_{\textcolor{blue}{+12.7}}$        \\
\midrule
Qwen3VL$_{2B}$  
& 20.0 & 17.3$_{\textcolor{red}{-2.7}}$  & 12.5 & 7.6$_{\textcolor{red}{-4.9}}$ & 7.5 & 4.6$_{\textcolor{red}{-2.9}}$ & 10.6 & 5.6$_{\textcolor{red}{-5.0}}$ \\
Qwen3VL$_{4B}$  
& 53.3 & 53.3$_{\textcolor{gray}{\pm 0.0}}$  & 18.4 & 26.1$_{\textcolor{blue}{+7.7}}$ & 18.1 & 24.8$_{\textcolor{blue}{+6.7}}$ & 22.4 & 23.0$_{\textcolor{blue}{+0.6}}$ \\
\quad $\hookrightarrow$ GRPO & 65.1$_{\textcolor{blue}{+11.8}}$  & 69.8$_{\textcolor{blue}{+16.5}}$  & 20.4$_{\textcolor{blue}{+2.0}}$ & 31.6$_{\textcolor{blue}{+13.2}}$ & 25.6$_{\textcolor{blue}{+7.5}}$  & 33.0$_{\textcolor{blue}{+11.9}}$ & 25.2$_{\textcolor{blue}{+2.8}}$ & 30.4$_{\textcolor{blue}{+8.0}}$ \\
Qwen3VL$_{8B}$  
& 59.0 & 57.6$_{\textcolor{red}{-1.4}}$  & 14.2 & 28.0$_{\textcolor{blue}{+13.8}}$ & 14.2 & 21.9$_{\textcolor{blue}{+7.7}}$ & 19.0 & 18.4$_{\textcolor{red}{-0.6}}$ \\
\quad $\hookrightarrow$ GRPO & 79.0$_{\textcolor{blue}{+20.0}}$  & 64.6$_{\textcolor{blue}{+5.6}}$  & 22.6$_{\textcolor{blue}{+8.4}}$ & 33.2$_{\textcolor{blue}{+19.0}}$ & 21.7$_{\textcolor{blue}{+7.5}}$  & 30.1$_{\textcolor{blue}{+15.9}}$ & 18.8$_{\textcolor{red}{-0.2}}$ & 28.9$_{\textcolor{blue}{+9.9}}$ \\
Qwen3VL$_{32B}$ 
& 62.2 & 60.6$_{\textcolor{red}{-1.6}}$  & 35.6  & 43.2$_{\textcolor{blue}{+7.6}}$ & 33.4  & 39.8$_{\textcolor{blue}{+6.4}}$ & 32.7 & 39.9$_{\textcolor{blue}{+7.7}}$ \\
Qwen3VL$_{235B}$ 
& 75.9 & 75.0$_{\textcolor{red}{-0.9}}$  & 24.5  & 47.3$_{\textcolor{blue}{+22.8}}$ & 35.4 & 50.2$_{\textcolor{blue}{+14.8}}$ & 35.1 & 48.9$_{\textcolor{blue}{+13.8}}$ \\
\hline 
InternVL$_{2B}$ & 45.9  & 40.8$_{\textcolor{red}{-5.1}}$  & 7.8 & 7.7$_{\textcolor{red}{-0.1}}$ & 3.1  & 3.5$_{\textcolor{blue}{+0.4}}$ & 3.7 & 4.1$_{\textcolor{blue}{+0.4}}$ \\
InternVL$_{4B}$ & 61.2  & 57.1$_{\textcolor{red}{-4.1}}$   & 8.3  & 21.0$_{\textcolor{blue}{+12.7}}$ & 5.0  & 16.3$_{\textcolor{blue}{+11.3}}$ & 5.5 & 15.2$_{\textcolor{blue}{+9.7}}$ \\
InternVL$_{8B}$ & 33.2 & 51.5$_{\textcolor{blue}{+18.3}}$  & 9.9  & 20.9$_{\textcolor{blue}{+11.0}}$ & 10.5  & 17.6$_{\textcolor{blue}{+7.1}}$ & 6.7  & 16.8$_{\textcolor{blue}{+10.1}}$ \\
InternVL$_{14B}$& 58.4 & 53.9$_{\textcolor{red}{-4.5}}$  & 15.3  & 28.0$_{\textcolor{blue}{+12.7}}$ & 16.3 & 23.9$_{\textcolor{blue}{+7.6}}$ & 14.2 & 23.6$_{\textcolor{blue}{+9.4}}$ \\
InternVL$_{38B}$ & 48.7  & 56.1$_{\textcolor{blue}{+7.4}}$  & 27.2  & 29.9$_{\textcolor{blue}{+2.7}}$ & 24.5  & 26.1$_{\textcolor{blue}{+1.6}}$ & 23.9  & 25.7$_{\textcolor{blue}{+1.8}}$\\
\hline
GPT\textendash 4o & 64.7  & 71.5$_{\textcolor{blue}{+10.5}}$  & 41.6  & 46.6$_{\textcolor{blue}{+5.0}}$ & 44.0  & 44.6$_{\textcolor{blue}{+0.6}}$ & 38.8 & 44.3$_{\textcolor{blue}{+4.5}}$ \\
GPT\textendash 5.4 & 75.6  & 77.3$_{\textcolor{blue}{+1.7}}$  & 35.6  & 38.2$_{\textcolor{blue}{+2.6}}$ & 38.4  & 35.7$_{\textcolor{red}{-2.7}}$ & 38.2 & 45.3$_{\textcolor{blue}{+7.1}}$ \\
Seed\textendash1.8 
& 86.1  & 86.8$_{\textcolor{blue}{+0.7}}$  & 56.8  & 62.6$_{\textcolor{blue}{+5.8}}$ & 61.0  & 64.2$_{\textcolor{blue}{+3.2}}$ & 67.0 & 68.5$_{\textcolor{blue}{+1.5}}$ \\
Seed\textendash2.0 
& 90.4  & 91.5$_{\textcolor{blue}{+1.1}}$  & 73.3  & 76.3$_{\textcolor{blue}{+3.0}}$ & 76.2  & 78.5$_{\textcolor{blue}{+2.3}}$ & 83.7 & 83.7$_{\textcolor{gray}{\pm 0.0}}$ \\
Gemini2.5\textendash Pro  
& 87.3 & 88.1$_{\textcolor{blue}{+0.8}}$  & 71.7  & 72.9$_{\textcolor{blue}{+1.2}}$ & 71.8  & 72.7$_{\textcolor{blue}{+0.9}}$ & 75.9 & 76.5$_{\textcolor{blue}{+0.6}}$ \\
Gemini3.1\textendash Pro  
& 87.7 & 85.9$_{\textcolor{red}{-1.8}}$  & 71.4  & 71.4$_{\textcolor{gray}{\pm 0.0}}$ & 73.2  & 72.8$_{\textcolor{red}{-0.4}}$ & 80.1 & 82.2$_{\textcolor{blue}{+2.1}}$ \\
\bottomrule
\end{tabular}%
}
\caption{Performance comparison across prompting paradigms and GRPO training. $F_{1d}$ and $F_{1b}$ denote direct and background-enhanced prompts, respectively. \textcolor{blue}{Blue}/\textcolor{red}{red} values indicate increases/decreases relative to the original model's $F_{1d}$. Detection (D) and Correction (C) are reported at Image (I) and Token (T) levels.} 
\label{tab:joint_ocr_f1_comparision}
\end{table*}

\section{Experiments}\label{sec:benchmark_results}

\subsection{Benchmarked Models}  
To comprehensively evaluating VLM's visual text understanding and correction capabilities, we select VLM's from two categories: open-source VLMs and closed-source VLMs. For open-source VLMs, we select three state-of-art model series, \textbf{Qwen2.5-VL} (3B, 7B, 32B, 72B, 235B\footnote{Qwen3-VL-235B-A22B is a Mixture-of-Experts (MoE) model. For brevity, it is referred to as Qwen3-VL-235B throughout this paper.}) \cite{qwen2025qwen25technicalreport}, \textbf{Qwen3-VL}(2B, 4B, 8B, 32B)~\cite{bai2025qwen3vltechnicalreport}, \textbf{InternVL3.5} (2B, 4B, 8B, 14B, 38B) \cite{wang2025internvl35advancingopensourcemultimodal}.
For closed-source baselines, we consider three widely adopted proprietary models: \textbf{GPT-4o}  \cite{openai2024gpt4technicalreport}, \textbf{GPT-5.4}(OpenAI), \textbf{Seed-1.8/2.0}(ByteDance) and \textbf{Gemini 2.5-pro/3.1-pro}(Google), which represent the world's leading flagship multimodal models.
For OCR+LLM pipeline, we select Qwen3(8B, 32B, 235B) and Qwen3VL(8B, 32B, 235B) for a clear comparison with corresponding VLMs.

\subsection{Results}
Our direct Correction prompt results, presented in Table \ref{tab:baseline_result_qwen_intern_gpt_new}, evaluate open-source VLMs on detection and correction tasks.Table~\ref{tab:joint_ocr_f1_comparision} presents the $F_1$ scores of Background Information Enhancement strategies, including a performance comparison against the direct prompting baseline as well as the results for specific models after GRPO training using MLVR.
\paragraph{Direct Correction Prompt}
Detection involves identifying errors at the image or token level, while correction requires generating the corrected output. According to Table~\ref{tab:baseline_result_qwen_intern_gpt_new}, performance is highest in I-D, with Seed-1.8 achieving 86.1 and Gemini2.5-Pro achieving 87.3 in $F_{1}$. For I-C, however, the scores drops to 61.0 and 71.8, respectively. The gap is bigger for other models, for Qwen3-VL-235B achieves 75.9 in I-D but only 35.4 in I-C, indicating that correction is more challenging than detection. Similarly, Qwen2.5VL-32B achieves 41.1 in token detection (T-D), but drops to 36.5 in token correction (T-C), suggesting difficulty in generating corrected tokens. Seed 1.8 and Gemini 2.5-Pro demonstrate impressive performance, which achieve 67.0 and 75.9 in T-C respectively, significantly outperforming the other open-source models evaluated in experiments.

\paragraph{Background Information Enhancement Prompt}
As illustrated in Table~\ref{tab:joint_ocr_f1_comparision}, background information enhancement prompt further improves results, especially for larger models. For small models like Qwen2.5-VL-3B, Qwen3VL-2B and InternVL-2B, The Background Information Prompt negatively impacts small-parameter models but substantially boosts larger models. This is due to the performance gap in visual comprehension, where the inferior visual processing of smaller models acts as a bottleneck that constrains their potential capabilities. Case studies are provided in Appendix~\ref{apdx:case_study_prompt}.

\paragraph{Targeted Model Training} 
The results of targeted model training method(GRPO with MLVR) are also shown in Table~\ref{tab:joint_ocr_f1_comparision}, demonstrating significant improvements across most metrics for all trained models. For Qwen2.5VL-3B, GRPO notably increases performance in all detection and correction tasks, with a +9.7 F1 score boost in Image-Level Detection (I-D), and +7.9 in Token-Level Correction (T-C). Qwen2.5VL-7B shows similar improvements, particularly in I-D (+31.6 F1 score), though Token-Level Detection (T-D) sees a slight decrease of -0.4. Overall, GRPO enhances detection and correction tasks, especially for larger models like Qwen2.5VL-7B and Qwen3VL-8B. Case studies and further quantitative analysis are provided in Appendix~\ref{apdx:case_study_grpo} and Appendix~\ref{apdx:comparison_grpo}, respectively.

\subsection{Analysis}
\begin{figure}[t!]
    \centering
    \includegraphics[width=\linewidth]{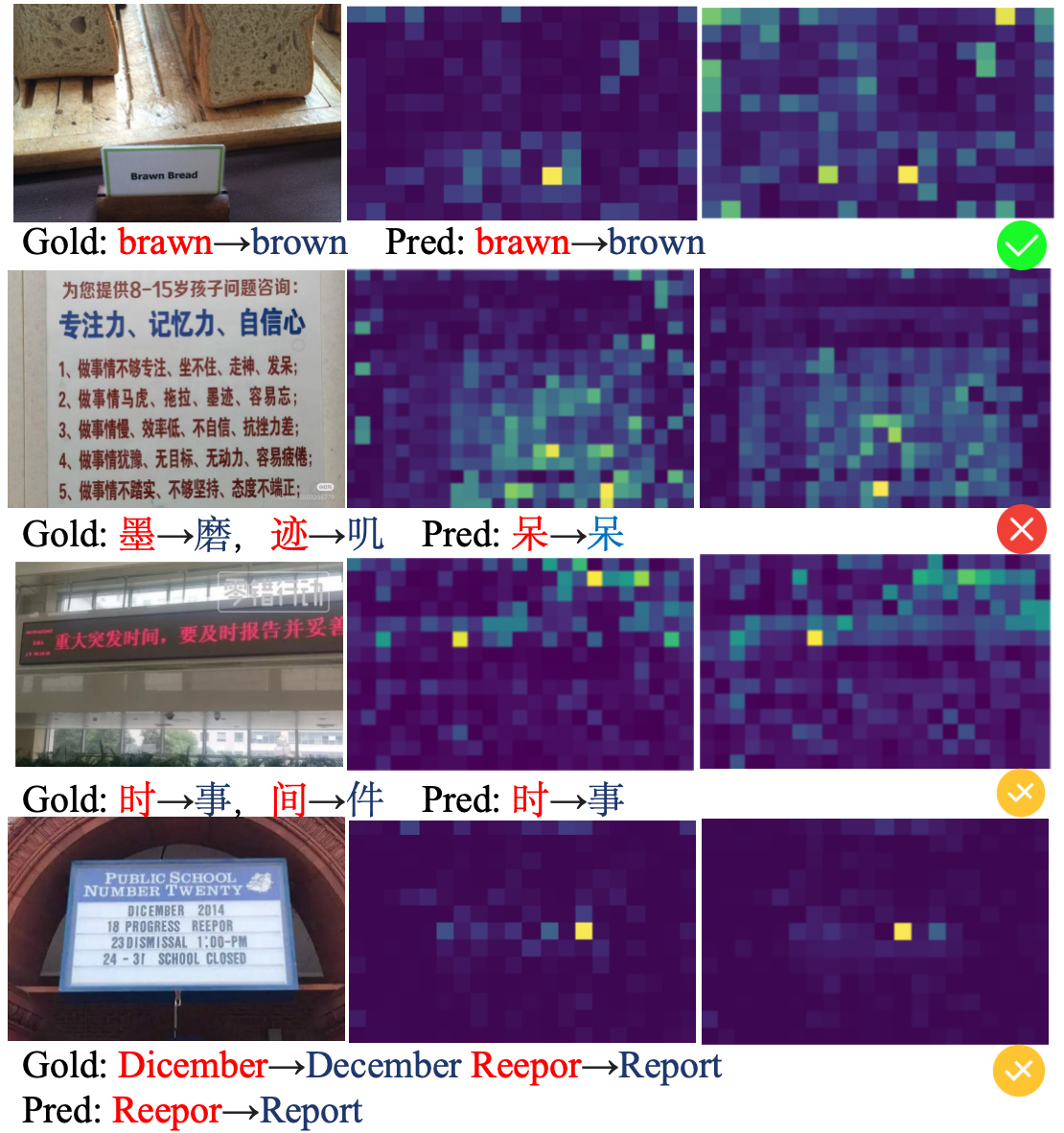}
    \caption{Case studies of attention visualization with QwenVL-7B.Each row shows the original image alongside attention maps from the 22nd and 25th layers. The corresponding golden labels and model predictions are displayed below, where incorrect tokens are highlighted in \textcolor{red}{red} and corrected tokens in \textcolor{blue}{blue}.}
    \label{fig:case_study}
    \vspace{-1em}
\end{figure}
\paragraph{Is this benchmark solvable by OCR+LLM pipeline?}
While smaller models like Qwen3-VL-8B benefit more from Oracle OCR text than from raw images, larger models encounter a bottleneck with text-only inputs. In contrast, deep integration of visual cues allows larger models to surpass this performance ceiling. Notably, Qwen2.5VL-32B and Qwen3VL-32B models achieve T-C scores of 36.5 and 32.7---both exceeding the best performance(31.7) attained using Oracle OCR inputs. These findings underscore the critical importance of a model’s ability to perform integrated reasoning over fine-grained visual details and textual content. Results for additional OCR tools, including specialized Models and VLMs are provided in Appendix~\ref{apdx:ocr_tools_llms}.

\paragraph{How do VLMs recognize error tokens in an image?}\label{para:attention_visualization_analysis} A case study in Figure~\ref{fig:case_study} demonstrates the attention mechanism of Qwen2.5-VL-7B. In Case 1, it successfully identifies the error, but in Case 2, the model fails to attend to the erroneous characters, even hallucinating the character "\chinese{呆}" not present in the image. Cases 3 and 4 show that the model can correct some errors but overlooks others, indicating difficulty in handling multiple errors simultaneously. We use the methodology from \citet{zhang2025mllmsknowlooktrainingfree} to visualize the attention maps. 

\paragraph{Does Model Scaling Necessarily Guarantee Performance Gains?}
Notably, smaller models like Qwen2.5-VL-3B (73.3) and Intern3.5-VL-4B (67.7) excel in I-D recall, likely due to over-correcting error-free images. In contrast, larger models like Qwen2.5-VL-32B (96.9) and Qwen3-VL-32B (97.3) achieve high precision by assuming images are error-free. Increasing the size of Qwen2.5-VL and Qwen3-VL does not guarantee better performance in all metrics. Qwen2.5-VL-72B shows a 3\% drop in T-C (33.1 vs. 36.5) compared to Qwen2.5-VL-32B. This is likely due to Qwen2.5-VL-32B's tendency to generate additional reasoning steps, improving correction accuracy, while Qwen2.5VL-72B strictly follows instructions. Conversely, in the InternVL3.5 family, an increase in model scale is positively correlated with stable gains in overall performance.

\begin{figure}[t] % [t] 建议图片放在页面顶部
    \centering
    \includegraphics[width=1.0\linewidth]{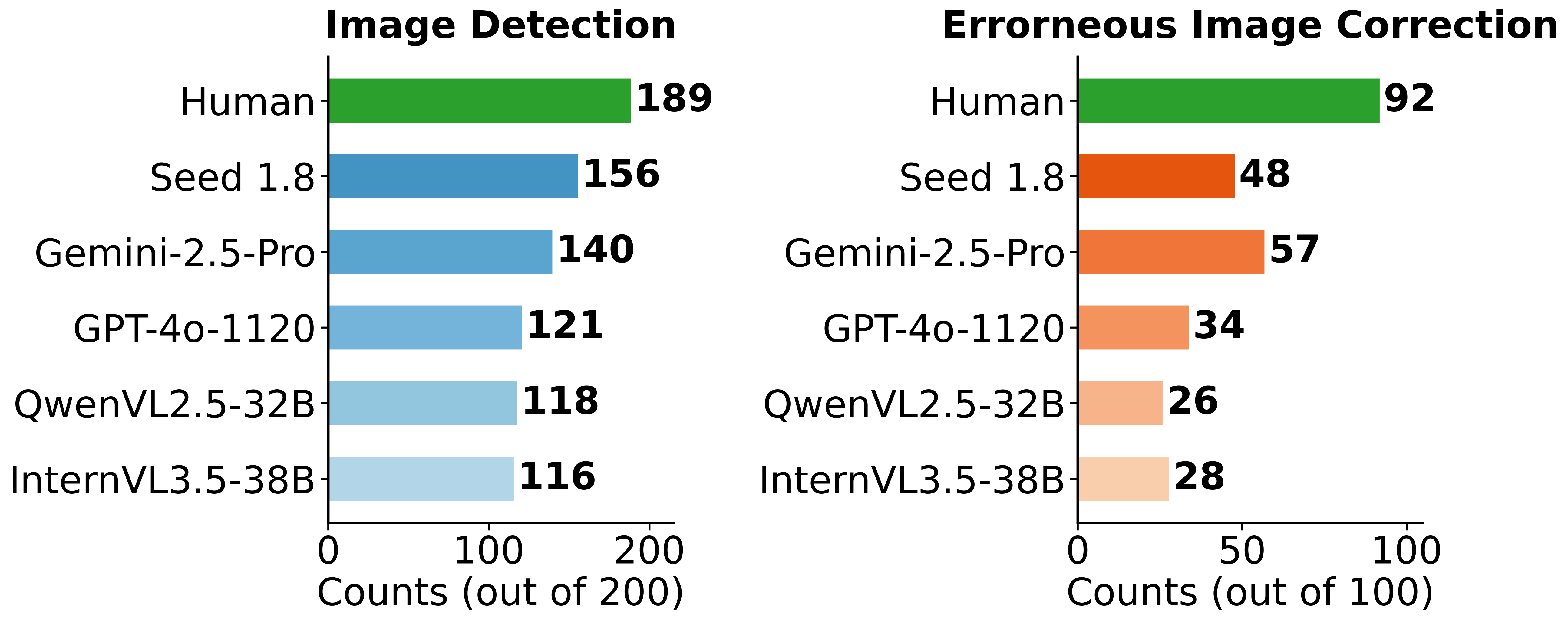}
    \caption{Performance of humans and VLMs on randomly selected images. Numbers indicate the counts of correctly classified images in image detection and fully corrected images in wrong image correction.}
    \label{fig:human_evaluation}
\end{figure}

\paragraph{How can humans detect and correct error tokens in an image?}

We randomly selected 200 images, equally split between Chinese and English, with each language containing 50 error-free and 50 erroneous samples. We evaluated two open-source models (Qwen2.5-VL-32B, Intern3.5-VL-38B) and three closed-source models (GPT-4o, Gemini 2.5-Pro, Seed-1.8) across two dimensions: (i) image-level detection and (ii) image-level correction. Two native speakers (one Chinese and one English) conducted the human evaluation. As shown in Figure~\ref{fig:human_evaluation}, humans significantly outperform VLMs in detection with a 94.5\% success rate. This gap becomes even more pronounced in correction: humans successfully rectified 82 out of 100 erroneous images, whereas the top-performing VLM, Gemini, managed 57, and the lowest-performing Qwen3-VL-32B achieved only 26. These results demonstrate a significant performance gap that still exists between current VLMs and human-level accuracy on this benchmark.

\section{Conclusion}
\label{sec:conslusion}
In this paper, we introduce ReViCo, a benchmark that evaluates VLMs through Visual Text Error Correction (VTEC). Success in VTEC demands that models transcend basic OCR to synthesize textual functions within complex visual contexts. Our evaluation reveals that despite gains from prompting and targeted training, a substantial performance gap remains between state-of-the-art VLMs and human-level reasoning. These findings underscore a persistent struggle in bridging visual cues with textual semantics. By exposing these limitations, ReViCo advocates for future research to prioritize the fine-grained integration of visual perception and linguistic reasoning to develop truly text-aware multimodal systems.

\section{Limitations}
While our benchmark and training methodologies are designed to be language-agnostic, their current validation is confined to English and Chinese datasets due to the availability of accessible data resources. Consequently, the generalizability of our approach to other languages, particularly low-resource ones, remains to be further explored. Additionally, the models evaluated are pre-trained VLMs, and performance could vary with task-specific fine-tuning. The focus on visual text errors excludes other linguistic challenges, such as grammatical or contextual mistakes. Lastly, while the proposed methods yield substantial improvements, a performance gap remains relative to human-level precision. This disparity highlights the ongoing challenges in fine-grained visual-textual perception and complex multimodal reasoning within current VLMs.

\section{Ethical Considerations}

\subsection{Legal Compliance, Copyright, and Privacy in Data Collection}
All data included in our benchmark was collected strictly following source terms of use and copyright policies, using tools widely employed in prior research. To protect privacy, all personally identifiable information and sensitive metadata were manually screened and removed. We ensure that the data is used solely for academic research purposes and that its distribution complies with the original licenses.

\subsection{Human Annotation}

The ReViCo test set contains 730 Chinese and 499 English instances collected from real-world images, including signs, menus, product labels, and other everyday scenes. Since annotation mainly involves identifying and correcting visual text errors in common scenarios, specialized linguistic expertise is not required. To ensure reliability, we recruited annotators through an internal university crowdsourcing platform. All held at least a Master's degree and had native or near-native proficiency in the annotated language, verified by qualifications such as IELTS, TOEFL, or GRE scores for English and the Putonghua Proficiency Test for Chinese.

Each instance was independently annotated by two annotators proficient in the corresponding language. Before annotation, they received detailed guidelines covering the task definition, annotation procedure, and criteria for identifying and correcting visual text errors. Annotation lasted three weeks, with an average workload of approximately 35 Chinese or 24 English instances per annotator per day. Project leads conducted regular quality checks to identify inconsistent labels and reduce individual bias.

The initial annotations achieved an image-level agreement rate of 92.6\%. Disagreements occurred in 91 instances, including 63 Chinese and 28 English cases, and were adjudicated by a third senior annotator. The resulting Cohen's $\kappa$ coefficient was 0.852, indicating strong agreement beyond chance and supporting the reliability of the guidelines and labels. All annotators were compensated at rates significantly above the applicable local minimum wage, reflecting the task's complexity and time requirements.

\section*{Acknowledgements}
We would like to express our sincere gratitude to the data annotators for their rigorous efforts in constructing the dataset and the anonymous reviewers for their insightful feedback and constructive suggestions.

\bibliography{custom}

\clearpage

\newpage
%\onecolumn
\appendix
\section*{Appendix}
\section{Error Type and Examples in ReViCo}
\label{apdx:error_type_explanation}

\begin{figure*}[h]
    \centering
    \includegraphics[width=\linewidth]{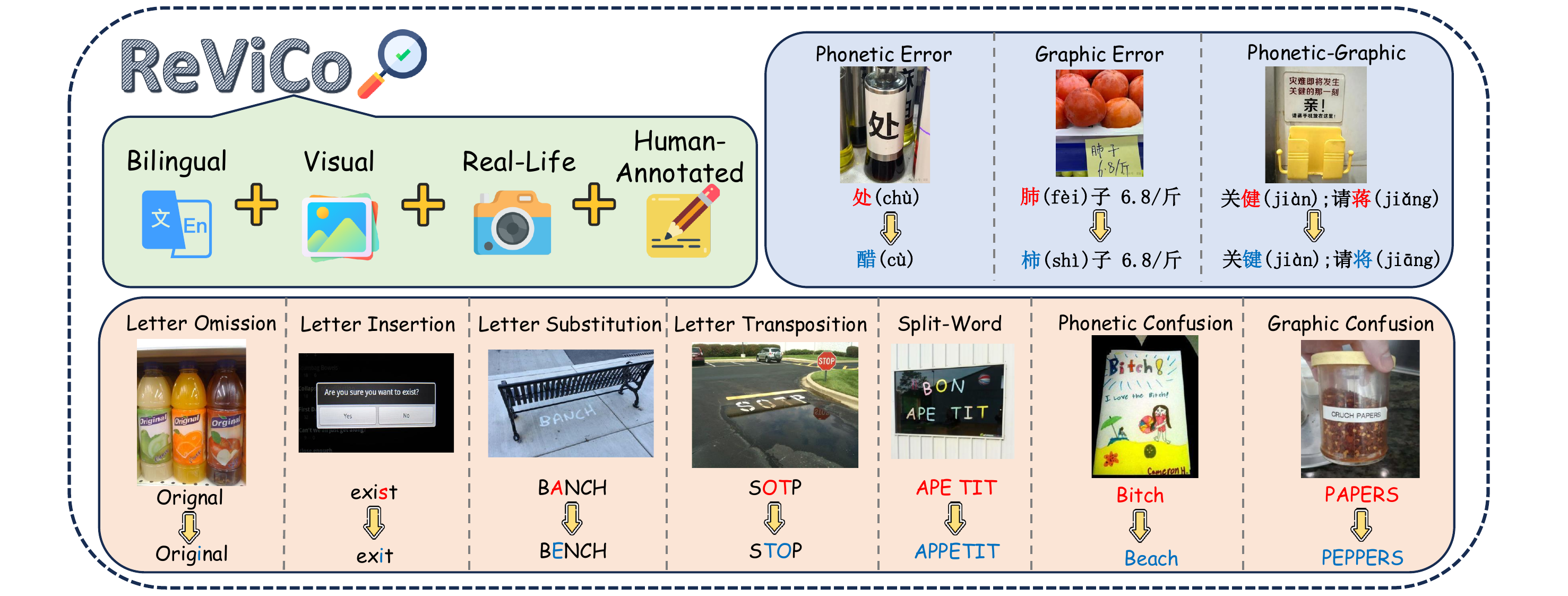}
    \caption{Examples from our ReViCo Benchmark, where the error types for Chinese and English are displayed. Wrong tokens are highlighted in \textcolor{red}{red} and the corresponding correct tokens are highlighted in \textcolor{blue}{blue}.}
    \label{fig:error_type_examples}
    \vspace{-1em}
\end{figure*}

Please find the error type at Table~\ref{tab:error_types} and Error examples in Figure~\ref{fig:error_type_examples}. 

\begin{table*}[h]
\centering
\small % 使用小号字体以适应较多内容
\begin{tabular}{p{0.18\linewidth} p{0.25\linewidth} p{0.5\linewidth}}
\toprule
\textbf{Language} & \textbf{Error Type} & \textbf{Explanation} \\
\midrule
\multirow{4}{*}{Chinese} & Phonetic Error & Substitution involving characters with similar pronunciation. \\
\cmidrule(lr){2-3}
& Graphic Error & Substitution involving characters with similar visual appearance. \\
\cmidrule(lr){2-3}
& Phonetic-Graphic Error & Substitution involving both phonetic and graphic similarity. \\
\cmidrule(lr){2-3}
& Others & Instances that do not fall into the preceding categories. \\
\midrule
\multirow{4}{*}{English (Letter Level)} & Letter Omission & A letter is missing from a word. \\
\cmidrule(lr){2-3}
& Letter Insertion & An extra letter is incorrectly added to a word. \\
\cmidrule(lr){2-3}
& Letter Substitution & A correct letter is replaced by an incorrect one. \\
\cmidrule(lr){2-3}
& Letter Transposition & Two adjacent letters are swapped. \\
\midrule
\multirow{3}{*}{English (Word Level)} & Split-word Error & A single word is incorrectly broken into two or more. \\
\cmidrule(lr){2-3}
& Phonetic/Graphic Confusion & A word is incorrectly used in place of another with a similar sound or spelling. \\
\cmidrule(lr){2-3}
& Others & Instances that do not fall into the preceding categories. \\
\bottomrule
\end{tabular}
\caption{Hierarchical categorization of text error types in the ReViCo benchmark.}
\label{tab:error_types}
\end{table*}

\section{Error Type Statistics of reinforcement learning data}
\label{apdx:error_type_rf_data}

We construct reinforcement learning training data using the three-stage pipeline described in Section~\ref{subsec:reinforcement_learning_data}. Error statistics for the Chinese and English datasets are reported in Tables~\ref{tab:chinese_rf_data_errors} and~\ref{tab:english_hf_data_errors}, respectively.

\begin{table}[h]
\centering
\small
\begin{tabular}{lc}
\toprule
Error Type & Count \\
\midrule
Phonetic              & 649  \\
Graphic               & 238  \\
Phonetic + Graphic    & 409  \\
% Character Insertion   & 144  \\
% Character Deletion    & 155  \\
Other Error           & 417  \\
\midrule
Total                 & 1713 \\
\bottomrule
\end{tabular}
\caption{Distribution of error types in the Chinese reinforcement learning data.}
\label{tab:chinese_rf_data_errors}
\end{table}

\begin{table}[h]
\centering
\small
\begin{tabular}{lc}
\toprule
Error Type & Count \\
\midrule
Letter Substitution   & 1043 \\
Letter Omission       & 652  \\
Letter Insertion      & 766  \\
Phonetic Confusion    & 348  \\
Split Word            & 9    \\
\midrule
Total                 & 2818 \\
\bottomrule
\end{tabular}
\caption{Distribution of error types in the English reinforcement learning data.}
\label{tab:english_hf_data_errors}
\end{table}

\section{Prompts for Chinese and English Visual Text Correction}
\label{apdx: prompt_english_chinese}

\begin{tcolorbox}[ breakable, colback=gray!8, colframe=gray!60!black, title=\textbf{Chinese Visual Text Correction}, fontupper=\small, ] 

    \chinese{你是一个严格遵循指令的中文视觉文本错误检查工具。请你检查图片中的中文文本错误，并执行以下操作:

​   1. 检查所有汉字，若存在文本错误（包括\textbf{拼写、用词不当、同音异形}等），则按照下面的格式输出错误的字及其对应的正确字:[("错误字","正确字"), ("错误字","正确字"), ...]；若未发现错误，则输出：["no error"]；

​   2. 请只关注中文文本的错误，不要关注语法错误和标点符号错误；

​   3. 请直接按照要求的格式输出结果，不要一步一步的思考分析。}

\end{tcolorbox}

\begin{tcolorbox}[ breakable, colback=gray!8, colframe=gray!60!black, title=\textbf{English Visual Text Correction}, fontupper=\small, ] 

    You are an English visual text correction tool. Please check the text in the image and follow these rules exactly:

​ 1. Check every word and if any error(including misspelling, misusing, homographies et. al.) exists, output in this exact format:[("error1","correction1"),("error2","correction2"),...], and if there is no error, output: ["no error"]

​ 2. Only check English word's error(ignore grammar/punctuation);

​ 3. Please directly output the result in the required format without thinking step by step.

\end{tcolorbox}

\section{Prompts for Background Information Enhancement}
See Table~\ref{tab:prompt_refinement} to get the prompts for Background Information Enhancement.
\label{apdx:prompt_improve}
\begin{table}[h]
\centering

\begin{tabular}{p{0.44\linewidth} p{0.44\linewidth}}
\hline
\multicolumn{2}{l}{\textbf{ Background Information Enhancement}} \\
\hline
(Chinese) \chinese{请先对图片的内容（包括文本、环境和背景信息等）进行概括} & 
(English) \textit{Please first summarize the content of the image (including text, environment and background information et al.)} \\
\hline
\end{tabular}
\caption{Additional prompts Background Information Enhancement, where the prompts are added to the direct prompt (Appendix~\ref{apdx: prompt_english_chinese}) as the first step.}
\label{tab:prompt_refinement}
\end{table}

\section{Calculation of Metrics}
\label{apdx:calculation_of_metrics}
For each image $i$, the gold annotation is a (possibly empty) set of error--correction pairs
$G_i=\{(w,c)\}$, and the model prediction is $P_i=\{(w,c)\}$.
Images without visual text errors have $G_i=\varnothing$, and predictions indicating no error are mapped to $P_i=\varnothing$.
English tokens are lower-cased before evaluation.
All metrics are micro-averaged with a small constant $\epsilon$ for numerical stability.

\paragraph{Metrics.}
Precision ($P$), recall ($R$), and $F_1$ are computed as
\begin{align*}
P &= \frac{TP}{TP+FP+\epsilon},\\
R &= \frac{TP}{TP+FN+\epsilon},\\
F_1 &= \frac{2TP}{2TP+FP+FN+\epsilon}.
\end{align*}
For image-level tasks, accuracy is
\[
\mathrm{Acc}=\frac{TP+TN}{TP+TN+FP+FN+\epsilon}.
\]

\paragraph{Image Detection (I-D)}
Image Detection is a binary decision of whether an image contains any visual text error:
\[
y_i=\mathbb{I}[G_i\neq\varnothing], \qquad \hat{y}_i=\mathbb{I}[P_i\neq\varnothing].
\]
$TP$, $FP$, $FN$, and $TN$ are accumulated over images based on $(y_i,\hat{y}_i)$.

\paragraph{Token Detection (T-D)}
Token Detection evaluates localization of erroneous tokens, ignoring corrections.
Let
\[
\begin{aligned}
E(G_i) &= \{\,w:(w,c)\in G_i\,\},\\
E(P_i) &= \{\,w:(w,c)\in P_i\,\}.
\end{aligned}
\]
Then
\begin{align*}
TP &= \sum_i |E(G_i)\cap E(P_i)|,\\
FN &= \sum_i |E(G_i)\setminus E(P_i)|,\\
FP &= \sum_i |E(P_i)\setminus E(G_i)|.
\end{align*}

\paragraph{Image Correction (I-C)}
Image Correction is counted as correct only if all errors are fixed exactly:
\begin{align*}
TP &= \sum_i \mathbb{I}[G_i\neq\varnothing \land P_i=G_i],\\
FN &= \sum_i \mathbb{I}[G_i\neq\varnothing \land P_i\neq G_i],\\
FP &= \sum_i \mathbb{I}[G_i=\varnothing \land P_i\neq\varnothing],\\
TN &= \sum_i \mathbb{I}[G_i=\varnothing \land P_i=\varnothing].
\end{align*}

\paragraph{Token Correction (T-C)}
Token Correction evaluates exact error--correction pairs:
\begin{align*}
TP &= \sum_i |G_i\cap P_i|,\\
FN &= \sum_i |G_i\setminus P_i|,\\
FP &= \sum_i |P_i\setminus G_i|.
\end{align*}

This strict setting penalizes both missed corrections and over-corrections.

\paragraph{Why four metrics?}
These four settings capture complementary aspects of visual text error correction.
Image-level detection (I-D) measures whether a model can recognize the existence of errors, while token-level detection (T-D) evaluates its ability to localize erroneous characters or words.
Token-level correction (T-C) assesses the correctness of individual error--correction pairs, independent of completeness.
Image-level correction (I-C) is the most stringent setting, requiring all errors in an image to be corrected exactly with no omissions or over-corrections.
Together, these metrics provide a comprehensive evaluation of error awareness, localization accuracy, correction quality, and holistic reliability.

\section{OCR tools + LLM Experiments Results}
\label{apdx:ocr_tools_llms}
We select professional OCR tools(GLM-OCR, HunyuanOCR) and powerful VLMs(Qwen2.5-VL-72B, Qwen3-VL-235B-A22B) to extract text, and test several LLMs as well as VLMs(used as LLM). Full results are presented in Table~\ref{tab:ocr_tools_llm_results}. From the results, we observe that using Oracle OCR Text consistently yields the best performance across all LLMs. In contrast, other OCR tools introduce cascading errors from imperfect recognition, which subsequently degrades correction performance. For instance, configurations using the HunyuanOCR model as the OCR engine exhibit the lowest I-C and T-C accuracies across nearly all tested models.

\begin{table*}[!t]
\centering
\scalebox{0.8}{
\begin{tabular}{lcccccc|cccccc}
\toprule
\multirow{2}{*}{Models} & \multicolumn{3}{c}{\textbf{I}mage-\textbf{D}etection} & \multicolumn{3}{c}{\textbf{T}oken-\textbf{D}etection} 
 & \multicolumn{3}{c}{\textbf{I}mage-\textbf{C}orrection} & \multicolumn{3}{c}{\textbf{T}oken-\textbf{C}orrection} \\ 
\cmidrule(lr){2-4} \cmidrule(lr){5-7} \cmidrule(lr){8-10} \cmidrule(lr){11-13}
 & $P$ & $R$ & $F_1$ & $P$ & $R$ & $F_1$ & $P$ & $R$ & $F_1$ & $P$ & $R$ & $F_1$ \\
\midrule
Qwen3$_{8B}$ 
& 70.7 & \underline{44.1} & \underline{54.3} & \underline{6.7} & \underline{27.5} & \underline{10.8} & \underline{30.3} & \underline{8.0} & \underline{12.6} & \underline{34.3} & \underline{15.4} & \underline{21.3} \\
\quad $\hookrightarrow$ GLM$--$OCR & 60.9 & 40.0 & 48.3 & 4.3 & 19.1 & 7.1 & 17.4 & 5.4 & 8.2 & 27.8 & 11.0 & 15.8 \\
\quad $\hookrightarrow$ HunyuanOCR & \underline{74.0} & 17.0 & 27.7 & 4.9 & 8.2 & 6.1 & 20.8 & 1.6 & 2.9 & 29.3 & 5.2 & 8.8 \\
\quad $\hookrightarrow$ Qwen2.5$--$VL$--$72B & 63.8 & 41.9 & 50.6 & 4.5 & 19.4 & 7.3 & 20.3 & 6.1 & 9.3 & 28.7 & 11.9 &16.8  \\
\quad $\hookrightarrow$ Qwen3$--$VL$--$235B & 63.4 & 42.0 & 50.5 & 3.9 & 19.6 & 6.5 & 19.0 & 5.7 & 8.8 & 28.4 & 11.8 & 16.7 \\
\midrule
Qwen3$_{32B}$
& 63.7 & \underline{57.0} & \underline{60.2} & 9.8 & \underline{25.7} & 14.2 & 31.4 & \underline{14.5} & 19.8 & \underline{36.7} & \underline{20.5} & \underline{26.3} \\
\quad $\hookrightarrow$ GLM$--$OCR & 65.7 & 54.3 & 59.5 & 10.1 & 24.4 & 14.3 & 33.6 & 14.3 & \underline{20.1} & 36.3 & 19.0 & 24.9 \\
\quad $\hookrightarrow$ HunyuanOCR & \underline{76.8} & 30.4 & 43.6 & 11.8 & 12.0 & 11.9 & \underline{38.5} & 5.8 & 10.0 & 31.3 & 9.2 & 14.2 \\
\quad $\hookrightarrow$ Qwen2.5$--$VL$--$72B & 64.4 & 50.6 & 56.6 & 11.8 & 20.0 & 14.8 & 31.1 & 12.6 & 18.0 & 33.4 & 15.9 & 21.5  \\
\quad $\hookrightarrow$ Qwen3$--$VL$--$235B & 65.7 & 53.1 & 58.7 & \underline{12.5} & 24.8 & \underline{16.6} & 33.2 & 13.1 & 18.8 & 35.4 & 19.8 & 25.4 \\
\midrule
Qwen3$_{235B}$
& 77.4 & \underline{58.6} & \textbf{66.7} & 24.4 & \textbf{33.2} & \textbf{28.1} & 55.1 & \textbf{21.0} & \textbf{30.4} & 45.6 & \underline{24.1} & \underline{30.2} \\
\quad $\hookrightarrow$ GLM$--$OCR & 66.5 & \underline{58.6} & 62.3 & 9.3 & 26.1 & 13.7 & 35.1 & 16.0 & 21.9 & 37.9 & 21.3 & 27.3 \\
\quad $\hookrightarrow$ HunyuanOCR & \underline{82.9} & 15.1 & 25.6 & 17.9 & 7.9 & 11.0 & \underline{56.5} & 4.0 & 7.6 & 43.1 & 6.6 & 11.5 \\
\quad $\hookrightarrow$ Qwen2.5$--$VL$--$72B & 77.3 & 43.5 & 55.6 & \underline{27.6} & 23.5 & 25.4 & 55.2 & 15.7 & 24.5 & \underline{47.5} & 19.7 & 27.8 \\
\quad $\hookrightarrow$ Qwen3$--$VL$--$235B & 77.3 & 47.2 & 58.6 & 15.9 & 25.2 & 19.5 & 53.6 & 16.0 & 24.7 & 45.1 & 20.6 & 28.3  \\
\midrule
Qwen3VL$_{8B}$ 
& 75.5 & 44.2 & 55.8 & \underline{12.3} & \underline{24.8} & \underline{16.5} & 44.5 & \underline{11.5} & 18.2 & \underline{38.9} & \underline{16.9} & \underline{23.6} \\
\quad $\hookrightarrow$ GLM$--$OCR & \underline{77.9} & \underline{47.7} & \underline{59.2} & 10.3 & 26.1 & 14.7 & \underline{47.2} & 12.1 & \underline{19.2} & 35.1 & 16.1 & 22.1 \\
\quad $\hookrightarrow$ HunyuanOCR & 72.6 & 16.5 & 26.9 & 4.7 & 6.4 &  5.4 & 33.3 & 3.1 & 5.7 & 35.2 & 5.8 & 10.0 \\
\quad $\hookrightarrow$ Qwen2.5$--$VL$--$72B & 77.6 & 44.3 & 56.4 & 10.3 & 21.0 & 13.8 & 45.6 & 10.5 & 17.1 & 31.9 & 13.3 & 18.7 \\
\quad $\hookrightarrow$ Qwen3$--$VL$--$235B & 77.1 & 47.2 & 58.6 & 9.9 & 24.4 & 14.0 & 45.1 & 11.4 & 18.2 & 35.4 & 16.1 & 22.1 \\
\midrule
Qwen3VL$_{32B}$ 
& 80.0 & \underline{39.7} & \underline{53.1} & 20.2 & \underline{21.9} & \underline{21.0} & 59.1 & \underline{14.3} & \underline{23.1} & \underline{49.0} & \underline{18.9} & \underline{27.3} \\
\quad $\hookrightarrow$ GLM$--$OCR & 80.4 & 38.5 & 52.1  & 17.7 & 20.7 & 19.1 & 58.3 & 13.1 & 21.5 & 42.1 & 15.3 & 22.4 \\
\quad $\hookrightarrow$ HunyuanOCR & \textbf{98.3} & 8.9 & 16.3 & \underline{60.3} & 6.5 & 11.8 & \textbf{96.4} & 4.2 & 8.1 & 62.0 & 5.5 & 10.2 \\
\quad $\hookrightarrow$ Qwen2.5$--$VL$--$72B & 79.9 & 34.8 & 48.5 & 14.5 & 18.0 & 16.1 & 58.5 & 12.3 & 20.4 & 43.7 & 14.2 & 21.4 \\
\quad $\hookrightarrow$ Qwen3$--$VL$--$235B & 80.8 & 37.1 & 50.9 & 14.4 & 20.3 & 16.8 & 59.1 & 12.7 & 21.0 & 46.1 & 16.6 & 24.4 \\
\midrule
Qwen3VL$_{235B}$ 
& 74.6 & \textbf{60.2} & \underline{66.6} & 15.4 & 28.5 & 20.0 & 49.6 & \underline{20.2} & \underline{28.7} & 43.3 & \textbf{25.8} & \textbf{31.7} \\
\quad $\hookrightarrow$ GLM$--$OCR & 72.8 & 53.7 & 61.8 & 17.4 & 23.9 & 20.1 & 46.0 & 17.1 & 25.0 & 40.2 & 20.2 & 26.9 \\
\quad $\hookrightarrow$ HunyuanOCR & \underline{88.5} & 14.3 & 24.7 & \textbf{31.6} & 9.9 & 15.0 & \underline{76.5} & 6.1 & 11.3 & \textbf{60.6} & 9.1 & 15.9 \\
\quad $\hookrightarrow$ Qwen2.5$--$VL$--$72B & 77.6 & 57.0 & 65.7 & 11.5 & \underline{30.1} & 16.6 & 46.2 & 14.1 & 21.6 & 37.0 & 20.5 & 26.4 \\
\quad $\hookrightarrow$ Qwen3$--$VL$--$235B & 74.8 & 57.8 & 65.2 & 19.1 & 27.0 & \underline{22.4} & 50.6 & 19.9 & 28.6 & 42.6 & 23.8 & 30.2 \\
\bottomrule
\end{tabular}
}
\caption{Performance comparison of OCR+LLMs with Oracle OCR Text and OCR tools on visual text error detection and correction, evaluated at both the image (I) level and token (T) level across detection (D) and correction (C) tasks. The best overall results in each column are highlighted in \textbf{bold}, while the strongest results within each LLM's results are \underline{underlined}. Please note that the first line of each LLM's results represents Oracle OCR Text+LLM.}
\label{tab:ocr_tools_llm_results}
\end{table*}

\section{Experiment Setting}
\paragraph{Open-source VLMs}
We deploy open-source VLMs on two NVIDIA A100 GPUs, using the default settings of each model. 
\paragraph{Closed-source VLMs}These closed-source systems complement the open-source VLMs, offering a comprehensive benchmark for evaluating spatial understanding across different accessibility settings. For closed-sourced VLMs, we use official APIs. The GPT-4o model is gpt-4o-2024-1120, the Seed-1.8 model is Seed-1.8-20251218 and the Gemini model is Gemini2.5-Pro released in mid 2025.

\label{apdx:exp_setting}

\section{Training Details}
We conducted GRPO training on Qwen2.5-VL-3B, Qwen2.5-VL-7B and Qwen3-VL-8B on four NVIDIA A100 GPUS, using our proposed Multi-Level Verification Reward (MLVR). Taking Qwen2.5-VL-7B as a primary case, the detailed training configurations are summarized in Table~\ref{tab: training details grpo}.
\begin{table}[H]
\small
\centering
\renewcommand{\arraystretch}{1.2}
\begin{tabular}{c|c}
\toprule
Hyper-parameters & Value \\ \hline
Training steps & 100 \\
Batch size & 512 \\
Warmup ratio & 0.03 \\
Learning rate scheduler & Cosine \\
Learning rate & $1\times10^{-6}$ \\
Rollout & 6 \\
KL Loss Coefficient & 0.01 \\
GPUs & 4 \\
Optimizer & Adam \\ \bottomrule
\end{tabular}
\caption{The hyper-parameters used during GRPO of Qwen2.5-VL-7B using the training set in ReViCo benchmark.}
\label{tab: training details grpo}
\end{table}

\section{Case Studies of Prompt-based Strategy}
\label{apdx:case_study_prompt}
\begin{figure*}[h]
    \centering
    \includegraphics[width=\linewidth]{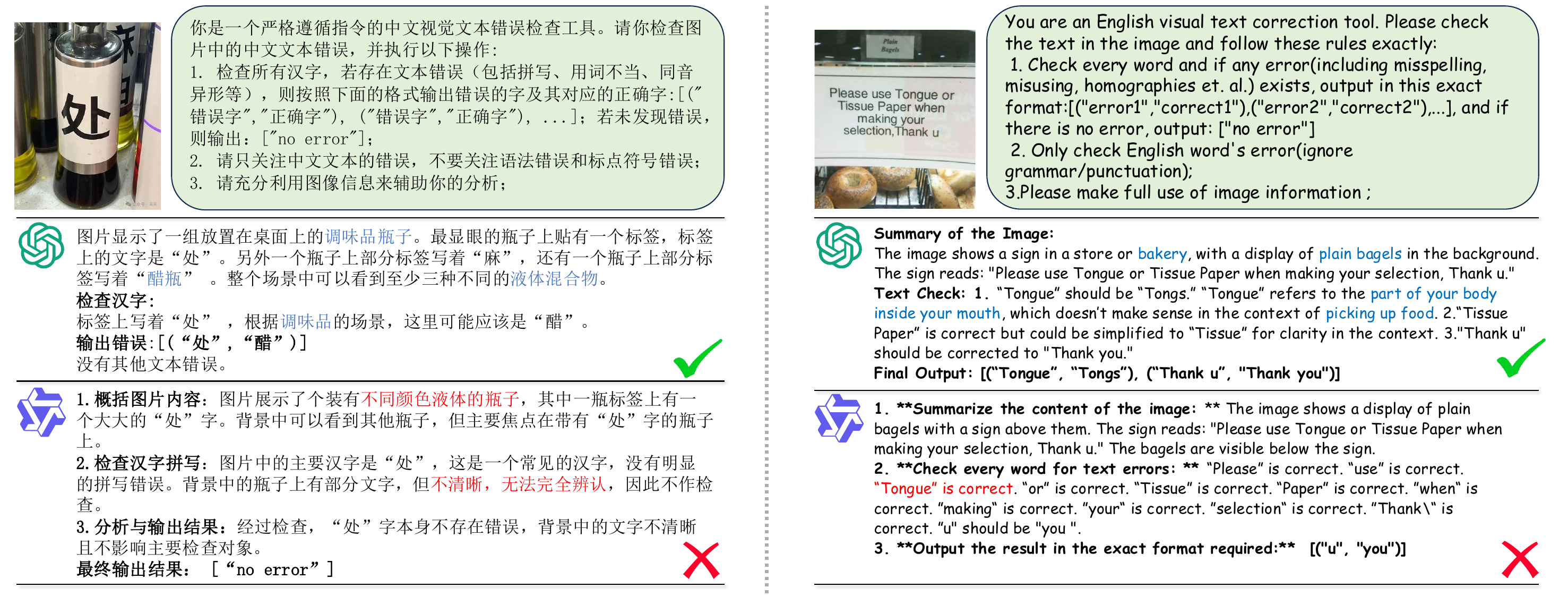}
    \caption{Examples of visual error correction by GPT-4o and QwenVL-32B under background information enhancement. The relevant background information is highlighted in \textcolor{blue}{blue}, while incorrect or vague descriptions are highlighted in \textcolor{red}{red}.
}
    \label{fig:bg_examples}
\end{figure*}

Case studies (Figure~\ref{fig:bg_examples}) highlight the value of background information. In one example, the Chinese character \textcolor{red}{\chinese{处}} (\pinyin{chù}, place) on a sticker should be corrected to \textcolor{blue}{\chinese{醋}} (\pinyin{cù}, vinegar), but QwenVL-32B fails to make the correct correction. The surrounding context—spice containers filled with black vinegar—provides a clear cue, yet the model misses this and does not correct the error. In contrast, GPT-4o successfully utilizes the visual context and background information, accurately identifying the correction. Similarly, the phrase "tongues" should be corrected to "tongs" based on a bakery sign, but QwenVL-32B struggles to capture the correct contextual meaning. These findings suggest that while background information can be highly beneficial, its utility depends on the model’s ability to accurately interpret and leverage it, emphasizing the need for strong reasoning capabilities in VLMs.

\section{Case studies of GRPO}
\label{apdx:case_study_grpo}
\begin{figure}[!t]
    \centering
    \includegraphics[width=\linewidth]{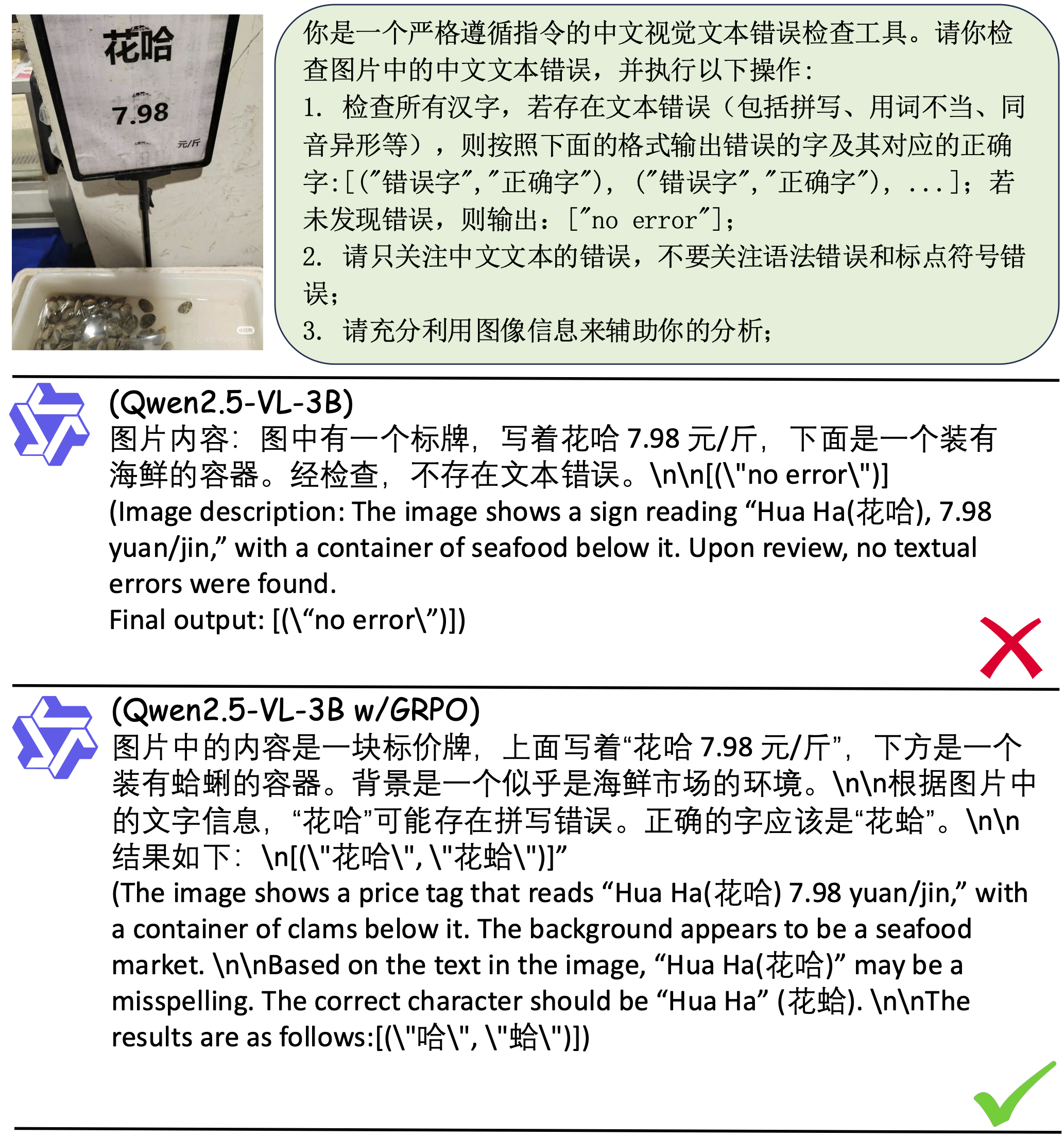}
    \caption{Chinese visual text error correction for Qwen2.5VL-3B without(up) and with targeted model training(bottom), original VLM response is in Chinese, English translation is given.}
    \label{fig:grpo_example}
\end{figure}

\begin{figure}[!t]
    \centering
    \includegraphics[width=\linewidth]{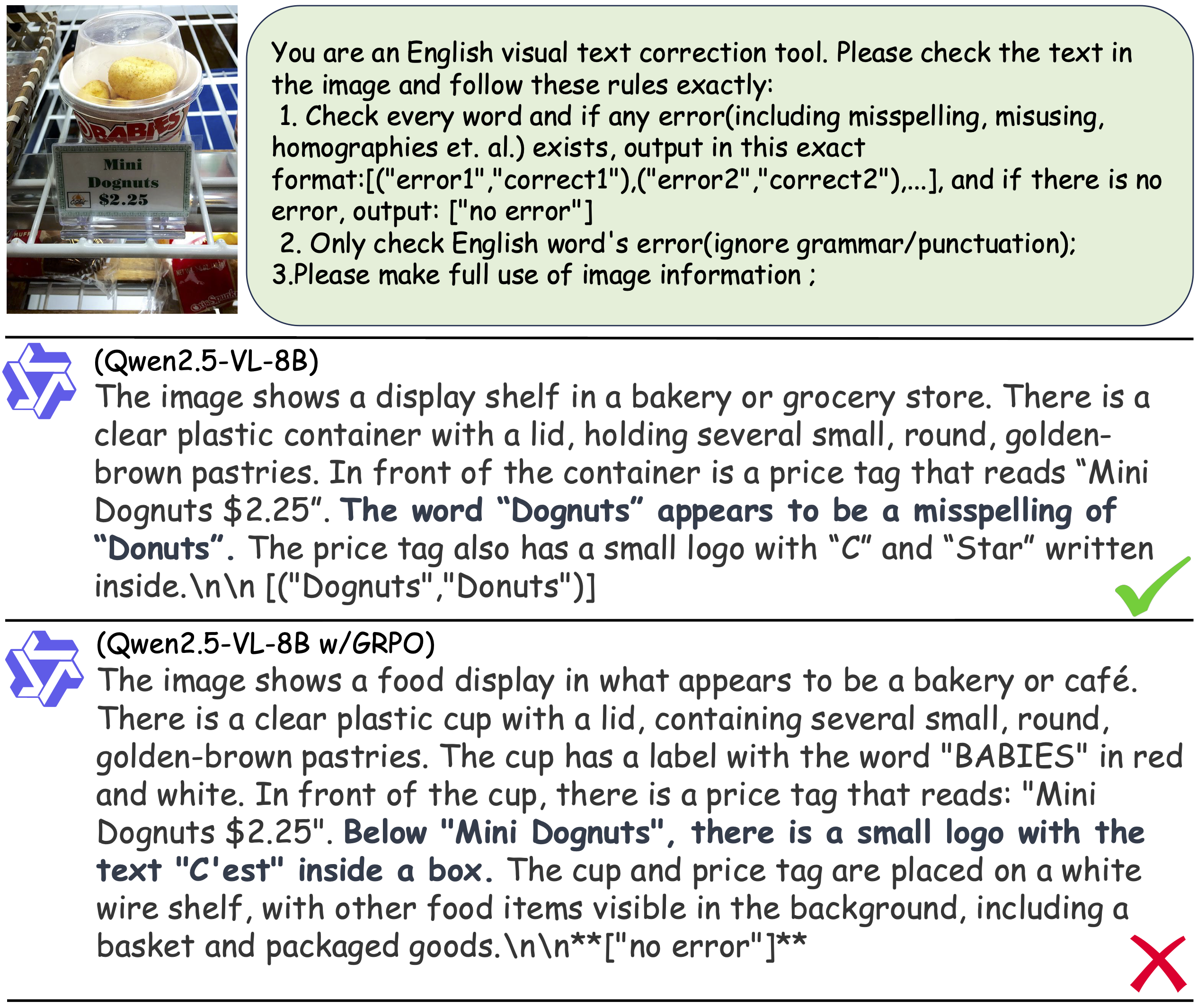}
    \caption{English visual text error correction for Qwen3VL-8B without(up) and with targeted model training(bottom).}
    \label{fig:grpo_example_fail}
\end{figure}

An example in Figure~\ref{fig:grpo_example} illustrates GRPO's effectiveness, where it successfully corrects a graphic error, changing "\chinese{花\textcolor{red}{哈}}" to "\chinese{花\textcolor{blue}{蛤}}". As illustrated in the figure, the model trained with GRPO effectively captures fine-grained visual details, enabling it to perform informed analysis and yield accurate results. Of course, GRPO doesn't always work. Figure~\ref{fig:grpo_example_fail} shows a failure case after GRPO training: the base Qwen3-VL-8B model was originally able to complete the text correction for this image correctly, but it actually failed after the training process. Comparing the two responses, it is evident that the GRPO-trained model does identify more accurate visual details, but it seems to get lost in the visual grounding process and just ignores the error text.

\begin{table*}[!htbp]
\centering
\setlength{\tabcolsep}{4pt}
\scalebox{0.9}{
\begin{tabular}{lcc|cc|cc|cc}
\toprule
\textbf{Model} 
& \multicolumn{2}{c}{\textbf{I-D}} 
& \multicolumn{2}{c}{\textbf{T-D}} 
& \multicolumn{2}{c}{\textbf{I-C}} 
& \multicolumn{2}{c}{\textbf{T-C}} \\
 & $F_{1d}$ & $F_{1b}$ 
 & $F_{1d}$  & $F_{1b}$ 
 & $F_{1d}$  & $F_{1b}$ 
 & $F_{1d}$  & $F_{1b}$ \\
\midrule
\multicolumn{8}{c}{\textit{Oracle OCR Text + Large language Models}} \\
\midrule
Qwen3VL$_{4B}$  
& 57.6 & -- & 14.4 & -- & 16.0 & -- & 21.2 & -- \\
\quad $\hookrightarrow$ GRPO & 59.3$_{\textcolor{blue}{+1.7}}$ & -- & 17.5$_{\textcolor{blue}{+3.1}}$  & -- & 18.1$_{\textcolor{blue}{+2.1}}$  & -- & 23.0$_{\textcolor{blue}{+1.8}}$  & -- \\

Qwen3VL$_{8B}$   
& 55.8 & -- & 16.5 & -- & 18.2 & -- & 23.6 & -- \\
\quad $\hookrightarrow$ GRPO & 60.2$_{\textcolor{blue}{+4.4}}$ & -- & 19.2$_{\textcolor{blue}{+2.7}}$  & -- & 19.6$_{\textcolor{blue}{+1.4}}$  & -- & 26.1$_{\textcolor{blue}{+2.5}}$  & -- \\

\midrule
\multicolumn{8}{c}{\textit{Image + Vision Language Models}} \\
\midrule

Qwen3VL$_{4B}$  
& 53.3 & 53.3  & 18.4 & 26.1 & 18.1 & 24.8 & 22.4 & 23.0 \\
\quad $\hookrightarrow$ GRPO & 65.1$_{\textcolor{blue}{+11.8}}$  & 69.8$_{\textcolor{blue}{+16.5}}$  & 20.4$_{\textcolor{blue}{+2.0}}$ & 31.6$_{\textcolor{blue}{+13.2}}$ & 25.6$_{\textcolor{blue}{+7.5}}$  & 33.0$_{\textcolor{blue}{+11.9}}$ & 25.2$_{\textcolor{blue}{+2.8}}$ & 30.4$_{\textcolor{blue}{+8.0}}$ \\

Qwen3VL$_{8B}$   
& 59.0  & 57.6  & 14.2  & 28.0 & 14.2  & 21.9 & 19.0  & 18.4 \\
\quad $\hookrightarrow$ GRPO & 79.0$_{\textcolor{blue}{+20.0}}$  & 64.6$_{\textcolor{blue}{+7.0}}$  & 22.6$_{\textcolor{blue}{+8.4}}$ & 33.2$_{\textcolor{blue}{+5.2}}$ & 21.7$_{\textcolor{blue}{+7.5}}$  & 30.1$_{\textcolor{blue}{+8.2}}$ & 18.8$_{\textcolor{red}{-0.2}}$ & 28.9$_{\textcolor{blue}{+10.5}}$ \\

\bottomrule
\end{tabular}
}
\caption{Performance comparison of Qwen3-VL-4B and Qwen3-VL-8B under OCR+LLM and Image+VLM paradigms. $F_{1d}$ and $F_{1b}$ denote direct and background-enhanced prompts, respectively. Results marked with $\hookrightarrow$ GRPO represent the model trained via GRPO. For the OCR+LLM pipeline, only $F_{1d}$ is reported as background enhancement is not applicable to pure text. \textcolor{blue}{Blue}/\textcolor{red}{red} values indicate increases/decreases relative to the base model's performance without GRPO. Thinking mode is activated for the OCR+LLM pipeline.}
\label{tab:comparison_grpo_two}
\end{table*}


\section{Performance Comparison of Models Before and After GRPO under OCR+LLM and Image+VLM Paradigms}
\label{apdx:comparison_grpo}
Table~\ref{tab:comparison_grpo_two} presents a performance comparison of models trained via GRPO(i.e., the Targeted Model Training described above) across two generation paradigms: OCR+LLM and Image+VLM. The results indicate that while GRPO training yields only marginal improvements in the OCR+LLM paradigm, it leads to a significant performance boost in the Image+VLM setting, particularly when the background enhancement prompt is employed. This disparity demonstrates that the GRPO-trained models achieve error correction primarily by interpreting fine-grained visual cues and visual text, rather than merely enhancing their pure-text correction capabilities.
\end{document}